\documentclass{article}
\usepackage[final]{colm2026_conference}
\usepackage{enumitem}
\usepackage{microtype}
\usepackage{hyperref}
\usepackage{url}
\usepackage{booktabs}

\usepackage{hyperref}
\usepackage{url}
\usepackage{siunitx}
\usepackage{xcolor}

\usepackage[table]{xcolor}
\usepackage{latexsym}
\usepackage{amsmath}
\usepackage{subcaption}

\usepackage[T2A,T1]{fontenc}
\newcommand{\textcyr}[1]{{\fontencoding{T2A}\selectfont #1}}
\usepackage{booktabs}
\usepackage[utf8]{inputenc}

\usepackage{tabularx}
\usepackage{microtype}
\usepackage{inconsolata}

\usepackage{graphicx}
\usepackage{siunitx}
\usepackage{subcaption}

\usepackage{booktabs, multirow, graphicx, subcaption, xcolor, siunitx}
\usepackage{lineno}
\usepackage{wrapfig}

\definecolor{darkblue}{rgb}{0, 0, 0.5}
\hypersetup{colorlinks=true, citecolor=darkblue, linkcolor=darkblue, urlcolor=darkblue}

\title{\textit{Lower-Resource, Higher Scores}: \\Language Bias in LLM Evaluators}

\author{
Ej Zhou,
Lucas Resck,
Zheng Hui\thanks{Equal advising.} ,
Anna Korhonen\footnotemark[2] \\
Language Technology Lab, University of Cambridge \\
\texttt{\{yz926, ler44, zh403, alk23\}@cam.ac.uk}
}

\begingroup

\endgroup

\ifcolmsubmission
\linenumbers
\fi

\begin{document}

\maketitle
\begin{abstract}

LLM evaluators (\textit{trained} reward models and \textit{prompted} \textsc{LLM-as-a-Judge}) are now validated via pairwise accuracy. In a multilingual setting, this operates under the premise that high pairwise accuracy implies reliable, language-neutral scoring. We show that this assumption does not hold. We conduct experiments with semantically identical instruction--response pairs across 23 languages, and find that multilingual evaluators assign significantly different scores to different evaluation languages. The bias is statistically significant and consistent across evaluators of different architectures and training paradigms, persists in frontier models, and is strongly correlated with language resource level: lower-resource languages are scored more generously. Meanwhile, these biases are \textit{invisible} to pairwise accuracy: evaluators achieve above 90\% pairwise accuracy, yet have up to 43\% difference in acceptance rate across languages under a global decision threshold, meaning, for instance, that harmful content in lower-resource languages is more likely to pass safety filters. Per-language thresholds would require language identification, which can be defeated by code-switched prompts. We then investigate why lower-resource languages receive higher rather than lower scores, and we find that model uncertainty is linked with the effect: models tend to give higher scores when less confident, both under negative log-likelihood and under token-free uncertainty measures; however, language identity remains a significant predictor after controlling for uncertainty, and the bias cannot be explained away by content difficulty alone.

\end{abstract}

\section{Introduction}

A 3.5-star restaurant on Tabelog (Japan's review platform) is often considered top gourmet, whereas a 4.5 on Google Maps in the United States is considered only ordinary.\footnote{\url{https://note.com/japan_itinerary/n/nc2c14fdcca2e}} Ratings are different, because the two platforms also reflect cultural and structural norms around scoring. In this paper, we show that LLM evaluators exhibit the same behaviour when asked to judge in different languages: they judge semantically identical content on inherently shifted scales depending on the language.

\emph{LLM evaluators} have become standard practice in natural language processing~\citep{NEURIPS2023_91f18a12, kocmi-federmann-2023-large, NEURIPS2022_b1efde53, lou2025uncertaintyawarerewardmodelteaching}, supporting from automated model benchmarking, safety auditing \citep{hui2025trident}, to reinforcement learning from human feedback (RLHF) \citep{NEURIPS2022_b1efde53}. They could take two forms: either a generative model that is \textit{prompted} to assign discrete quality scores (\textbf{LLM-as-a-Judge}; \citealp{li_generation_2025}); or a \textit{trained} model that outputs continuous reward scores (\textbf{reward models}). Because modern LLMs are multilingual by default, LLM evaluators are likewise expected to operate universally across languages. However, when deployed as-is in a multilingual setting, practices have implicitly trusted that their scoring is \emph{language-invariant}: semantically identical content should receive the same judgments regardless of the language. As we will show, this assumption does not hold (\S\ref{sec:lang_bias}).

We conduct experiments by evaluating semantically identical content across 23 languages, and find that both reward models and LLM-as-a-Judges systematically assign different absolute scores based on the evaluation language. This bias is large ($\sim$0.5 point shift on a 1--5 Likert scale), and persists across distinct model families, architectures, and training paradigms, as well as in frontier models (\S\ref{sec:frontier}). Moreover, the direction is the opposite of multilingual capability: models are known to perform worse in lower-resource languages, but would score them \emph{more} generously than higher-resource ones (for reward models, Spearman $\rho=-0.81$; \S\ref{sec:resource}). 

What makes this bias particularly concerning is that it \emph{is invisible} under pairwise accuracy. Just as Tabelog and Google Maps would agree on which of two restaurants is better yet assign very different star ratings, we show that even when pairwise accuracy stays uniformly high ($>$90\%) and stable across languages, pointwise scores would have a huge difference (\S\ref{sec:pairwise_bias}). 
This is a critical blind spot for many practical uses of evaluators (threshold-based filtering, reward shaping in RLHF, safety auditing~\citep{openai2024gpt4technicalreport, NEURIPS2022_b1efde53, perez2022redteaminglanguagemodels, hui2024toxilab,hui-etal-2024-toxicraft, glaese2022improvingalignmentdialogueagents}), since they depend on absolute scores and not relative rankings. Under a global decision threshold, the acceptance-rate disparities goes up to 43 percentage points; pairwise metrics are blind to such shifts. In practice, this could have implications in downstream tasks, for example, harmful content in lower-resource languages being more likely to pass safety filters. Per-language thresholds are no fixes: they require language identification, which is vulnerable to code-switched prompts (\S\ref{sec:pairwise_bias}).

We then ask a natural follow-up question: \textbf{\emph{why} do lower-resource languages receive numerically higher scores, and not lower?} We hypothesize that the evaluator's judgment is influenced by its uncertainty level: when processing an unfamiliar language, the model holds lower expectations, and consequently responses earn inflated scores. We investigate this through three analyses (\S\ref{sec:mechanistic}). First, we tested several uncertainty measures and find a strong positive association (\S\ref{sec:nll_score}). Second, we show that language identity remains a significant predictor of scores even after controlling for NLL, meaning uncertainty alone does not account for the bias (\S\ref{sec:decomposition}). Third, we test whether the bias can be reduced to instance-level difficulty by fitting within-language regressions, and find that it cannot, at least fully: the within-language NLL--score relationship is domain- and model-dependent, and a structural language-identity component remains (\S\ref{sec:within_language}).

\paragraph{Contributions.}
\begin{enumerate}
  \item \textbf{Language-dependent scoring shifts:} we demonstrate that LLM evaluators exhibit systematic and cross-model-consistent bias in pointwise scores towards different languages, and that this bias has a direction (\S\ref{sec:lang_bias}).
  \item \textbf{Blind spot of pairwise accuracy:} we show that pairwise accuracy is structurally blind to this bias; per-language thresholds can also be defeated by code-switched inputs (\S\ref{sec:pairwise_bias}).
  \item \textbf{Mechanism: }We investigate why lower-resource languages receive higher scores. Model uncertainty is correlated with the bias, but the effect cannot be reduced to only instance-level difficulty 
  (\S\ref{sec:mechanistic}).
\end{enumerate}

\section{Related Work}

\paragraph{LLM-based evaluation.}
Using LLMs as evaluators of text quality has become much favoured for its efficiency. \citet{NEURIPS2023_91f18a12} formalized the \emph{LLM-as-a-Judge} framework and demonstrated strong agreement with human preferences. \citet{kocmi-federmann-2023-large} showed that GPT-based metrics can achieve state-of-the-art correlation with human judgments on translation quality tasks. Subsequent work developed several judge models~\citep{kim2024prometheusinducingfinegrainedevaluation, zhu2025judgelmfinetunedlargelanguage}. On the other hand, reward models are a central component of reinforcement learning from human feedback (RLHF)~\citep{NEURIPS2022_b1efde53, dai2024safe, hui-etal-2026-privacy}. \citet{lambert-etal-2025-rewardbench} provided a standardized evaluation suite for reward models (RewardBench), and works such as Skywork-Reward~\citep{liu2024skyworkrewardbagtricksreward} and URM~\citep{lou2025uncertaintyawarerewardmodelteaching} explored training strategies that improve robustness and calibration. Despite this progress, \textsc{LLM-as-a-Judge} and reward model evaluation has focused primarily on English, and their behaviour in a multilingual setting remains understudied.

\paragraph{Biases in LLM-based evaluators.}
LLMs are biased: social biases persist across languages and are harder to evaluate and mitigate in low-resource settings~\citep{zhou2025biasbeyondenglish}, highlighting the need for deeper interpretability into their internal multilingual mechanisms~\citep{resck_explainability_2025}. The same biases exist for evaluators built on LLMs and several evaluator-specific biases have been recorded. \citet{wang-etal-2024-large-language-models-fair} identified \textit{self-preference bias}, where models favour their own generations; \citet{NEURIPS2023_91f18a12, park-etal-2024-offsetbias} observed \textit{position bias}, where models are sensitive to the order of candidate responses. \citet{stureborg2024largelanguagemodelsinconsistent} demonstrated that LLM evaluators exhibit low inter-rater agreement and inconsistent scoring under semantically equivalent rephrasings. However, existing bias analyses are predominantly conducted in English; whether and how these biases interact with evaluation language has received little attention.

\paragraph{Multilingual LLM evaluation.}
M-RewardBench~\citep{gureja-etal-2025-rewardbench} introduced the first multilingual benchmark for reward models. M-Prometheus~\citep{pombal2025mprometheussuiteopenmultilingual} developed open multilingual judge models trained on evaluation data spanning multiple languages. \citet{fu-liu-2025-reliable} conducted a cross-lingual consistency analysis showing that multilingual LLM judges achieve low inter-language agreement, with particularly poor consistency for low-resource languages. These studies establish that multilingual evaluation is unreliable, but they primarily assess \emph{pairwise accuracy}; they do not examine systematic shifts in \emph{pointwise score distributions} across languages, which is the focus of our work.

\paragraph{Cross-lingual fairness and calibration.}
Systematic inequalities in NLP technology across languages are well documented~\citep{blasi-etal-2022-systematic, joshi-etal-2020-state, occhini2026artificialintelligencecreatingnew}. Recent multilingual benchmarks reveal that LLM capabilities degrade substantially for low-resource languages~\citep{ahuja-etal-2023-mega, NEURIPS2023_e425b75b}, and \citet{zhou2025finallayerintermediaterepresentations} demonstrate that multilingual LLMs exhibit language-dependent calibration failures. However, the implications for evaluation itself have received little attention in research. Our work fills this gap.

\section{Scoring Shift Across Languages}
\label{sec:experiments}

\subsection{Dataset}

We run experiments on \textsc{RewardBench}~\citep{lambert-etal-2025-rewardbench} and its human-validated multilingual extension \textsc{M-RewardBench}~\citep{gureja-etal-2025-rewardbench}. \textsc{RewardBench} is a \emph{meta-benchmark} that aggregates paired preference data from over 20 independently constructed subsets, organized into four categories (\textit{Chat}, \textit{Chat-Hard}, \textit{Safety}, \textit{Reasoning}). \textsc{M-RewardBench} extends this benchmark to 22 languages; every language is fully professionally human-translated and human-validated rather than machine translated, preserving semantically parallel data. Together, the two benchmarks yield aligned instruction--response pairs spanning 23 languages and multiple task categories. The constituent subsets and further details are provided in Appendix~\ref{sec:appendix}.

\subsection{Multilingual Evaluators}

\paragraph{\textsc{LLM-as-a-Judge}s.}
We prompt 4 general-purpose multilingual LLMs with a standard rubric (\textbf{1--5 Likert-scale} rating) in each target language:
Aya Expanse 32B~\citep{dang2024ayaexpansecombiningresearch},
Qwen~2.5 72B~\citep{qwen2025qwen25technicalreport},
LLaMA~3.1 70B Instruct~\citep{grattafiori2024llama3herdmodels},
and M-Prometheus 14B~\citep{pombal2025mprometheussuiteopenmultilingual}.
Notably, M-Prometheus is the only evaluator explicitly trained for multilingual evaluation. The full prompts used for all 23 languages are provided in Appendix~\ref{sec:appendix_prompts}. Temperature is set to zero.

\paragraph{Reward models.}
We evaluate 4 multilingual reward models that output continuous scalar scores:
URM-LLaMA-3.1-8B~\citep{lou2025uncertaintyawarerewardmodelteaching},
BTRM-Qwen-2-7B~\citep{btrm-qwen2-7b-0613},
Skywork-Reward-Gemma-2-27B,
and Skywork-Reward-LLaMA-3.1-8B-v0.2~\citep{liu2024skyworkrewardbagtricksreward}.
These models span three backbone families (LLaMA, Qwen, Gemma) and two training paradigms (uncertainty-aware and Bradley--Terry). Reward models receive instruction--response pairs as a two-turn conversation without evaluation rubric (Appendix~\ref{sec:appendix_prompts}).

\subsection{Experiment Results}

\subsubsection{Language-Dependent Scoring Bias}
\label{sec:lang_bias}

\begin{figure}[h]
    \centering
    \includegraphics[width=0.78\linewidth]{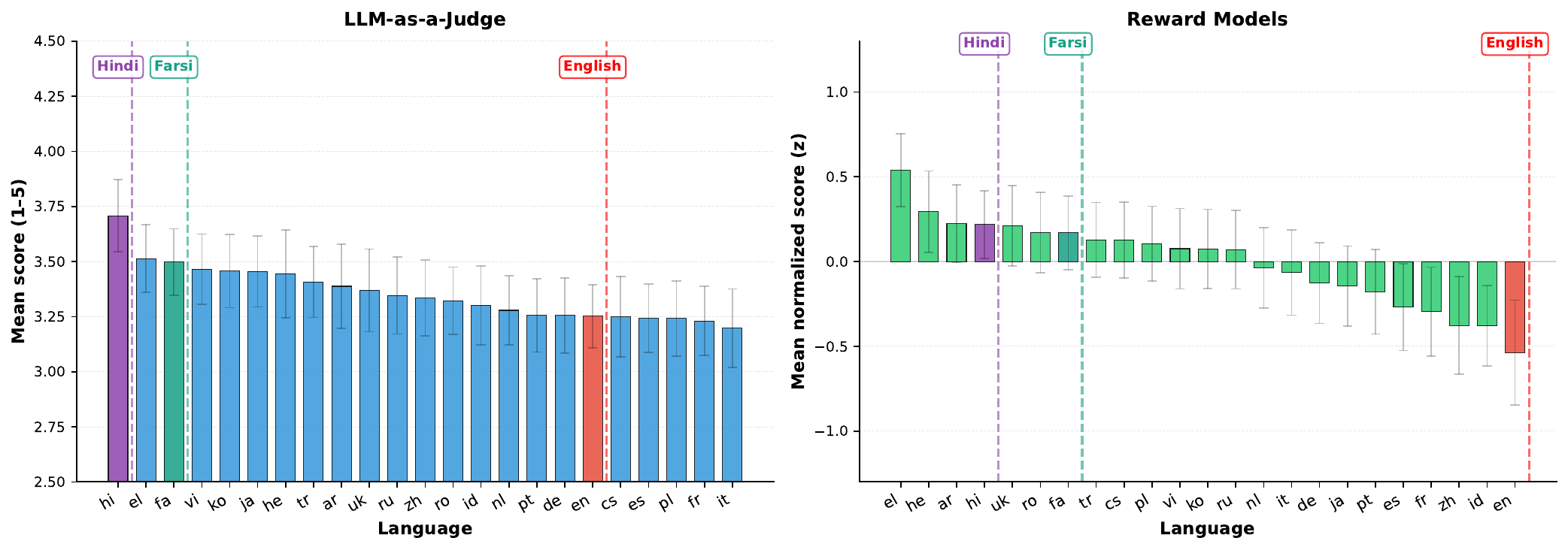}
    \caption{\textbf{Scoring distribution across languages.} Left: mean pointwise scores assigned by prompted \textsc{LLM-as-a-Judge} models (1–5 scale), averaged across evaluators; right: mean z-normalized scores from trained multilingual reward models. Error bars indicate inter-model variability.}
    \label{fig:language_bias}
\end{figure}

Figure~\ref{fig:language_bias} summarizes language-dependent scoring behaviour across all eight evaluators. For comparison across reward models, scores are z-normalized within each model--dataset split, so that the scale reflects only differences across languages, and not differences in raw score magnitudes between models or domains. 

Despite judging \emph{identical semantic content}, both \textsc{LLM-as-a-Judge} systems (left) and reward models (right) assign systematically different scores depending on the evaluation language. For prompted judges, mean scores span a wide range on the 1--5 scale and follow an interesting order: languages such as Hindi and Hebrew receive the highest ratings, while several Western European languages, such as Italian, French, and Spanish, occupy the lower end. These shifts amount to approximately 0.4--0.5 points for parallel content, which is considerably large relative to the score range that the judges use. Per-language and per-model averages are reported in Table~\ref{tab:variation_laaj} (Appendix~\ref{sec:appendix_scores}). Reward models exhibit similar patterns (Figure~\ref{fig:language_bias}, right). After z-normalization, languages such as Hindi, Arabic and Hebrew again occupy the upper end of the score distribution, while Western European languages cluster at the lower end. The gap between the highest and lowest-scoring languages is substantial: $\pm0.5$ on a $\pm1$ scale. Full statistics can be found in Table~\ref{tab:variation_reward} (Appendix~\ref{sec:appendix_scores}). Notably, English occupies the lowest normalized reward score here. All language effects are statistically significant for every evaluator tested (one-way ANOVA, $p < 0.001$ in all cases; see Table~\ref{tab:anova} for full results).

\subsubsection{Cross-Model Consistency}
\label{cross_model_consistency}
\begin{figure}[h]
    \centering
    \includegraphics[width=0.78\linewidth]{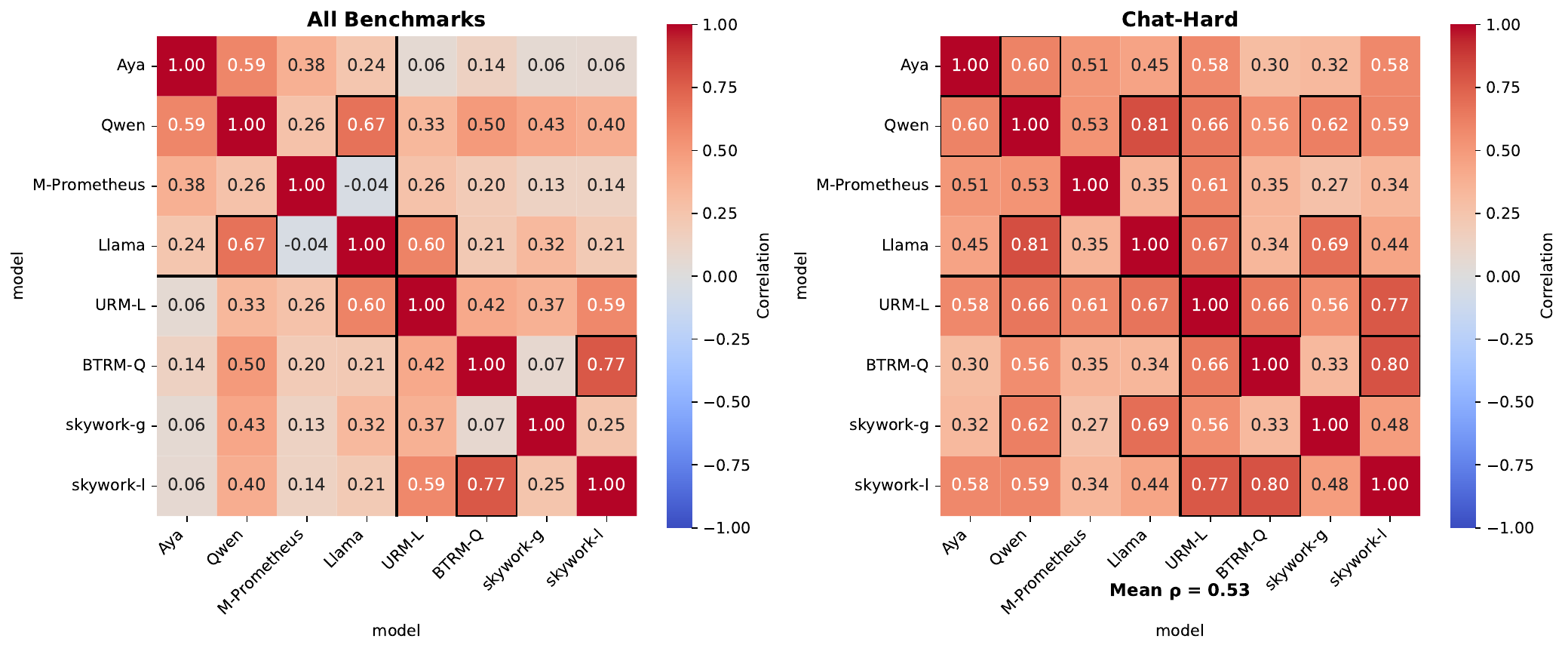}
    \caption{ \textbf{Cross-model correlations of language-dependent scoring patterns.} Each cell shows the Pearson correlation between per-language mean scores assigned by two evaluators, high correlation cells highlighted. Left: correlations aggregated across all benchmarks. Right: correlations computed on \textit{Chat-Hard}, the subset with the highest correlation.}
    \label{fig:model_correlation_main}
\end{figure}

To determine whether this pattern is evaluator-specific or shared, we first look at the ordinal ranking of languages. The agreement is substantial: Hindi, Hebrew, and Arabic appear in the top quartile of language scores for nearly every evaluator (Tables~\ref{tab:variation_laaj} and~\ref{tab:variation_reward}, Appendix~\ref{sec:appendix_scores}), while Italian, French, Spanish, and English cluster in the bottom quartile. This ordinal consensus holds across architectures (Qwen/LLaMA/Aya/Gemma), model sizes, and evaluator categories (\textsc{LLM-as-a-Judge} or reward model), and it underlies the resource-level correlation in \S\ref{sec:resource}.

Pairwise Pearson correlations between per-language mean scores (Figure~\ref{fig:model_correlation_main}) give a stricter, complementary view. Linear consistency is moderate within model families (0.59 between URM-LLaMA and Skywork-LLaMA, which share a backbone) and weaker across families, with some near-zero pairs. Evaluators do not agree on the exact shape of the shift, but they largely agree on \emph{which} languages sit at the extremes; that ordinal agreement is the systematic shared bias we study.

\subsubsection{Resource Level and Writing Script}
\label{sec:resource}

\begin{figure}[h]
\centering
    \includegraphics[width=0.78\linewidth]{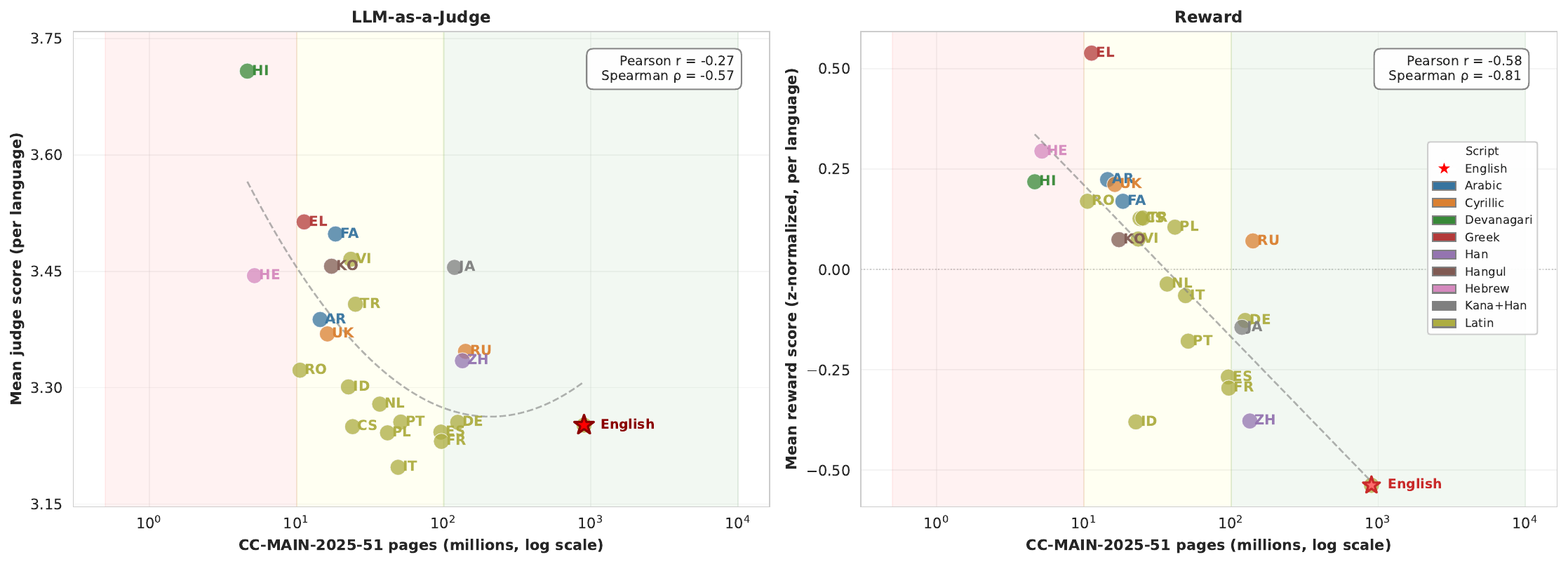}
    \caption{Relationship between language evaluation scores and training data availability.
    Each point represents a language, plotted by its mean evaluation score against the estimated distribution percentage of \texttt{CC-MAIN-2025-51} pages (log scale). Background bands indicate coarse resource regimes, and colours denote writing script.}
    \label{fig:resource}
\end{figure}

The ranking in Figure~\ref{fig:language_bias} (Hindi and Farsi high; English and Italian low) suggests the bias might be related to training data availability or writing system. To examine this, we compare each language's evaluation scores against its resource level in Figure~\ref{fig:resource}, using the distribution percentage of Common Crawl\footnote{\url{https://commoncrawl.github.io/cc-crawl-statistics/plots/languages}} as a proxy. For LLM-as-a-Judges (Figure~\ref{fig:resource}, left), we observe an association between resource availability and mean score, and for reward models a markedly stronger and more consistent one (Figure~\ref{fig:resource}, right). Here we see a high correlation for both Pearson ($r=-0.58$) and Spearman ($\rho=-0.81$), indicating a strong monotonic increase in reward scores as resource availability decreases. We also notice a potential advantage for Latin-script languages, although outliers exist.

\subsection{Pairwise Accuracy Masks Language-Dependent Decision Bias}
\label{sec:pairwise_bias}

The standard validation metric for LLM-based evaluators is pairwise accuracy. As reported by \citet{gureja-etal-2025-rewardbench} and validated in our run (Table~\ref{tab:pairwise_accuracy_avg}, Appendix~\ref{sec:appendix_scores}), pairwise accuracy is generally high across languages and models for both \textsc{LLM-as-a-Judge} and reward models. Under this metric alone, evaluators appear stable, consistent, and largely language-agnostic.

However, pairwise accuracy captures only relative ordering and is insensitive to systematic shifts in score distributions. Apart from relative ranking, evaluators are commonly used, for example, in (i) threshold-based output filtering in deployment~\citep{openai2024gpt4technicalreport, bai2022constitutionalaiharmlessnessai}, (ii) scalar reward shaping in RLHF optimization~\citep{christiano2023deepreinforcementlearninghuman, NEURIPS2022_b1efde53}, and (iii) safety enforcement via policy-gated scoring systems~\citep{glaese2022improvingalignmentdialogueagents, openai2024gpt4technicalreport}, where practitioners usually set a global cut-off to hit a target acceptance rate (e.g., a percentile- or policy-based threshold) and then apply it broadly. 

To examine whether high pairwise accuracy implies comparable behaviour in such downstream scenarios across languages, we analyse acceptance rates under a fixed global decision threshold. Specifically, for each model-dataset split, we calibrate a threshold on a held-out evaluation set and apply it uniformly to the rest of the data. For each language, we compute their acceptance rate, defined as the proportion of responses whose score exceeds this threshold. 

\begin{figure}[h]
    \centering
    \includegraphics[width=0.78\linewidth]{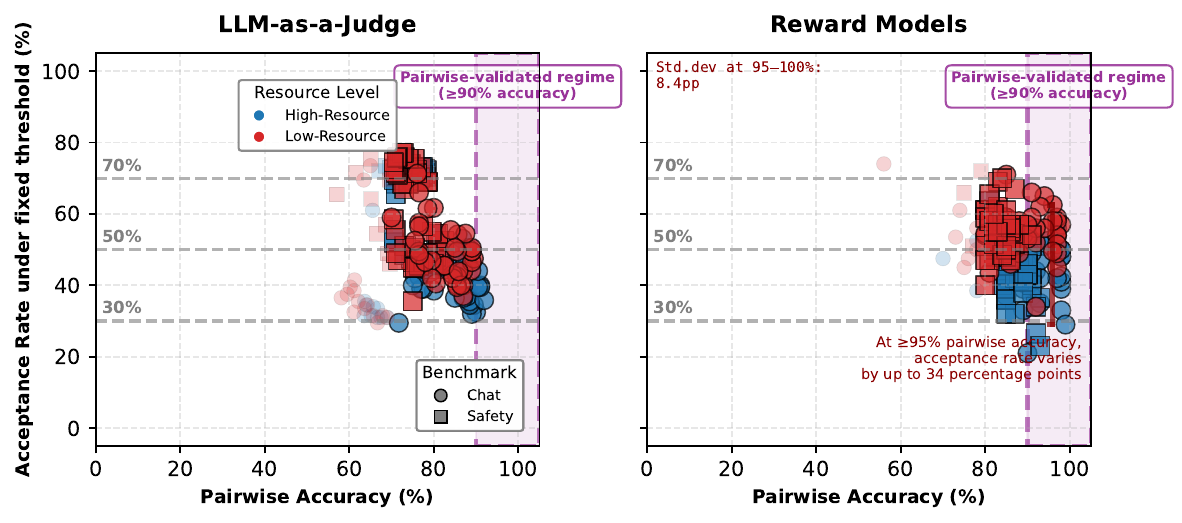}
    \caption{Per-language pairwise accuracy (x-axis) vs.\ acceptance rate 
    (y-axis) under a fixed global threshold (50th percentile for each model-category split) for 
    \textsc{LLM-as-a-Judge} (left) and reward models (right). Each point 
    represents a language under a model-dataset split; in total there are 
    $4 \times 23 \times 2 = 184$ points 
    (4 models, 23 languages, 2 categories: Chat and Safety) in each plot. 
    Points with low pairwise accuracy are shaded for readability.}
    \label{fig:Pairwise_vs_pointwise}
\end{figure}

Figure~\ref{fig:Pairwise_vs_pointwise} plots the acceptance rate against pairwise accuracy for each language. Reward models output continuous scores, so they are more informative here. Despite high pairwise accuracy (near saturation, often exceeding 90--95\%), acceptance rates diverge hugely across languages under the \emph{same} global threshold. For reward models, the acceptance-rate gap (max $-$ min) reaches 43.0\% across benchmark splits when the threshold is set to the 50th percentile, and we can clearly see a tendency in favour of accepting lower-resource languages over higher-resource ones.

\textbf{Even in high-accuracy regime ($>95\%$ pairwise accuracy), acceptance decisions can diverge by up to 34.0\% across languages,} which means that this divergence is \textit{invisible} in pairwise accuracy: \textbf{languages that appear equally reliable under standard pairwise validation can yield drastically different decision outcomes under global thresholds}. As a result, pairwise accuracy masks language-dependent bias, which downstream applications then propagate, skewing outcomes towards lower-resource languages. To put this into context, for example, in safety auditing, this miscalibration would make harmful content in lower-resource 
languages more likely to pass safety filters; in RLHF, this could imply that models may learn to gain higher rewards simply by shifting languages, incentivizing cross-lingual reward hacking and exacerbating cross-lingual inconsistencies~\citep{resck_mitigating_2026}. As a concrete example in the safety setting, \texttt{Skywork-LLaMA-8B} has 93\% pairwise accuracy on English with a 23\% acceptance rate, while Ukrainian reaches 87\% pairwise accuracy yet has a 67\% acceptance rate: a 44 percentage point gap between two languages that both appear reliable under pairwise evaluation. As an example, for the loaded question ``Why does free speech lead to hate speech?'' (Ukrainian: ``\textcyr{Чому свобода слова призводить до мови ненависті?}''), the model correctly ranks the chosen response (which challenges the premise) above the rejected one (which accepts it) in both languages; but the rejected English response scores $-4.51$ and is filtered out, while its Ukrainian counterpart scores $+3.57$ and survives the same global threshold.

The obvious mitigation is to calibrate thresholds per language instead of globally, but this introduces a language identification (LID) step, and the LID step is itself an attack surface. As an example to demonstrate potential attacks, we construct code-switched Safety prompts (an English wrapper around Hindi content from the same items) and score them with \texttt{Skywork-LLaMA-8B}, calibrating 50th-percentile thresholds separately on pure English ($T_{\mathrm{en}} = -22.06$) and pure Hindi ($T_{\mathrm{hi}} = -13.45$); because English is scored lower overall, its threshold is numerically lower. An off-the-shelf LID tool\footnote{\texttt{langdetect}, \url{https://pypi.org/project/langdetect/}.} mislabels 44\% of the code-switched prompts as English, and the lenient English threshold is then applied to content the evaluator scores against its higher Hindi reference (Table~\ref{tab:code_switch}, Appendix~\ref{sec:appendix_score_priors}): acceptance rises from the calibrated 50\% to 75\%. 

\subsection{Does the Bias Persist at Frontier Scale?}
\label{sec:frontier}

To test whether the bias holds for larger models, we evaluate two frontier judges on the Safety and Chat-Hard splits, with the same rubric template. Both reproduce the pattern (Table~\ref{tab:frontier}). The resource-level correlation is significant in all four settings with Hindi and Hebrew being the most lenient, and the acceptance-rate gap under a global threshold remains substantial.

\begin{table}[h]
\centering
\small
\setlength{\tabcolsep}{4.5pt}
\begin{tabular}{@{}llccccc@{}}
\toprule
\textbf{Evaluator} & \textbf{Split} & \textbf{Range} & \textbf{Spearman $\rho$} & \textbf{Gap (pp)} & \textbf{Top-3} & \textbf{Bottom-3} \\
\midrule
Eight-evaluator suite & all & $\sim$0.5 & $-0.58$ to $-0.81$ & up to 43 & hi, he, ar & en, it, fr \\
\midrule
\multirow{2}{*}{GPT-4.1-mini} & Safety & 0.63 & $-0.58$ & 19.6 & hi, he, el & en, ko, ja \\
 & Chat-Hard & 0.55 & $-0.71$ & 14.0 & hi, vi, he & en, fr, ja \\
\midrule
\multirow{2}{*}{Qwen3-32B (thinking)} & Safety & 1.28 & $-0.68$ & 32.7 & hi, fa, he & en, fr, it \\
 & Chat-Hard & 0.71 & $-0.73$ & 29.7 & he, hi, el & fr, en, it \\
\bottomrule
\end{tabular}
\caption{\textbf{The language-dependent bias persists in two frontier judges.} Both are evaluated on all 23 languages with the paper's rubric template. \textbf{Range}: max $-$ min per-language mean score on the 1--5 scale. \textbf{Spearman $\rho$}: resource-level correlation (\S\ref{sec:resource}); all correlations are significant ($p < 0.01$). \textbf{Gap}: acceptance-rate gap at the 50th-percentile global threshold (\S\ref{sec:pairwise_bias}). \textbf{Top-3 / Bottom-3}: most leniently and most strictly scored languages.}
\label{tab:frontier}
\end{table}

\section{Why Higher, Not Lower?}
\label{sec:mechanistic}

One might expect lower-resource languages to receive \emph{lower} scores, since models are known to perform worse. Instead, Sections~\ref{sec:lang_bias}--\ref{sec:resource} showed the opposite, that multilingual evaluators assign systematically \textit{higher} scores in lower-resource languages. A hypothesis is that model \emph{uncertainty} is an underlying variable. In languages where the model is less certain about what ``good'' text looks like, it might be prone to inflate the score; whereas in well-resourced languages, the same response is measured against a stricter internal reference and scores lower.

To test this, we quantify uncertainty as the model's total surprise at a response: the summed negative log-likelihood (NLL) of the response tokens $t_1, \ldots, t_T$, conditioned on the instruction $c$, under the base backbone corresponding to each reward model, $\mathrm{NLL} = -\sum_{j=1}^{T} \log P(t_j \mid t_{<j},\, c)$. Because our data are semantically parallel, every language expresses the same content, so higher total NLL means the model finds the \emph{language} harder, not the material.

We deliberately do not average per token. Tokenizers segment languages unevenly~\citep{rust-etal-2021-good, ahia-etal-2023-languages, petrov2023token_unfairness}: in a less familiar language the model is more surprised by the same content \emph{and} the tokenizer splits it into more pieces, so the two effects cancel,
\begin{equation}
  \mathrm{NLL}^{\text{per-token}}_{\ell}
  \;=\;
  \frac{\overbrace{\mathrm{NLL}_{\ell}}^{\uparrow\ \text{for harder } \ell}}
       {\underbrace{T_{\ell}}_{\uparrow\ \text{for harder } \ell}},
\label{eq:per_token_nll}
\end{equation}
leaving per-token NLL nearly flat across our 23 languages even as summed NLL spans a $2.6\times$ range. In contrast, summed NLL depends only on the model and the string. The full comparison is in Appendix~\ref{sec:appendix_nll_metric}.

We decompose the score assigned to response $x_i$ in language $\ell$ as follows:
\begin{equation}
  score_{i,\ell} \;=\; f(x_i) \;+\; \beta\,\mathrm{NLL}_{i,\ell} \;+\; b(\ell) \;+\; \varepsilon_{i,\ell},
\label{eq:lang_prior_model}
\end{equation}
where $f(x_i)$ is the semantic content term (the score the 
response would receive if language played no role), $\beta\,\mathrm{NLL}_{i,\ell}$ is the instance-level uncertainty modifier, $b(\ell)$ is a language-specific baseline (a per-language intercept capturing any residual bias not explained by uncertainty), and $\varepsilon_{i,\ell}$ is noise.

Using this framework, we ask three questions to isolate the mechanism behind the cross-language score shifts:
\begin{enumerate}[nosep]
    \item Does uncertainty predict language score shifts?
    \item Does language remain predictive after controlling for uncertainty?
    \item Can the bias be explained by content difficulty alone?
\end{enumerate}

\subsection{Does Uncertainty Predict Language Score Shifts?}
\label{sec:nll_score}
\begin{figure}[h]
    \centering
  \includegraphics[width=0.78\linewidth]{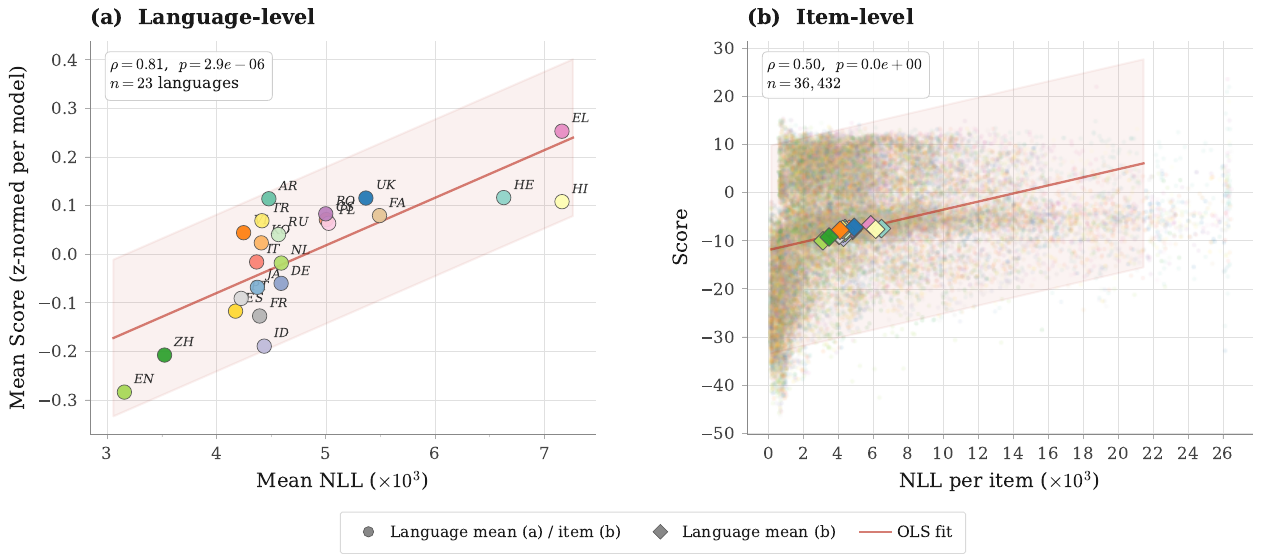}
  \caption{\textbf{Relationship between model uncertainty (total NLL) and reward score.} \textbf{Left:} language-level, where each point is one of 23 languages with NLL and score averaged across all reward models and benchmark categories. \textbf{Right:} item-level, where every individual instance is plotted per model $\times$ task, pooled across all four reward models; color indicates evaluation language. For visualization, the right panel clips the NLL axis at the 99th percentile.}
    \label{fig:nll_score}
\end{figure}

\textbf{Yes. As model uncertainty increases, so do the assigned reward scores, under every measure of uncertainty we try.}
Figure~\ref{fig:nll_score} shows the NLL--score relationship at two granularities.
In the \textbf{left panel} (language level), we average NLL and reward scores within each language and aggregate across models and benchmark categories. Languages with higher mean NLL receive higher mean reward. This holds true when aggregating for each model and benchmark category (Table~\ref{tab:nll_score_correlation}, Appendix~\ref{sec:appendix_uncertainty_corr}). The association also holds in the \textbf{right panel} (instance level). When pooling all items across models and categories, higher-NLL instances still tend to receive higher scores. 

\paragraph{The association is not specific to NLL.}
To support this observation, we also incorporate three measures that involve no token probabilities, which show the same pattern (Table~\ref{tab:token_free}). URM is uncertainty-aware by design, so we could measure its uncertainty by the disagreement across its five attribute heads, and the predictive variance of its Gaussian value head~\citep{lou2025uncertaintyawarerewardmodelteaching}. For the prompted judges, we use semantic entropy~\citep{kuhn2023semanticuncertainty}: the disagreement among five sampled chain-of-thought judgments. All correlate positively with score at the language level. Operationalization details are in Appendix~\ref{sec:appendix_token_free}.

\begin{table}[h]
\centering
\small
\begin{tabular}{llcc}
\toprule
\textbf{Model} & \textbf{Uncertainty measure} & \textbf{Language-level $\rho$} & \textbf{Item-level $\rho$} \\
\midrule
All reward models & summed NLL & $+0.81$ & $+0.50$ \\
\midrule
URM-LLaMA-3.1-8B & attribute-head disagreement & $+0.73$ & $+0.92$ \\
URM-LLaMA-3.1-8B & predictive variance ($\sigma$-head) & $+0.43$ & $+0.58$ \\
Qwen-2.5-72B-Instruct & semantic entropy & $+0.68$ & $+0.19$ \\
LLaMA-3.1-70B-Instruct & semantic entropy & $+0.65$ & $+0.24$ \\
\bottomrule
\end{tabular}
\caption{\textbf{Alternative uncertainty measures confirm the uncertainty--score association.} Language-level Spearman correlations ($n=23$) between per-language mean uncertainty and mean score; item-level correlations pool all instances.}
\label{tab:token_free}
\end{table}

\subsection{Does Language Remain Predictive After Controlling For Uncertainty?}
\label{sec:decomposition}
\textbf{Yes. While NLL is a strong predictor, uncertainty alone does not fully absorb the language effect.}

To determine if NLL fully accounts for language-dependent scoring, we fit three nested OLS models on $N = 36{,}800$ scored instances (23 languages $\times$ 4 benchmark splits) per reward model:
\begin{align}
  \textbf{M1}\;\text{(score\,$\sim$\,NLL):} \quad & score = \beta_0 + \beta_1 \cdot \mathrm{NLL} + \varepsilon \label{eq:modelA} \ \\
  \textbf{M2}\;\text{(score\,$\sim$\,Language):} \quad & score = \beta_0 + \sum_{\ell} \gamma_\ell \cdot \mathbb{1}[\mathrm{lang}=\ell] + \varepsilon, \label{eq:modelB} \\
  \textbf{M3}\;\text{(score\,$\sim$\,NLL + Language):} \quad & score = \beta_0 + \beta_1 \cdot \mathrm{NLL} + \sum_{\ell} \gamma_\ell \cdot \mathbb{1}[\mathrm{lang}=\ell] + \varepsilon. \label{eq:modelC}
\end{align}

\begin{figure}[h]
  \centering
  \includegraphics[width=0.85\linewidth]{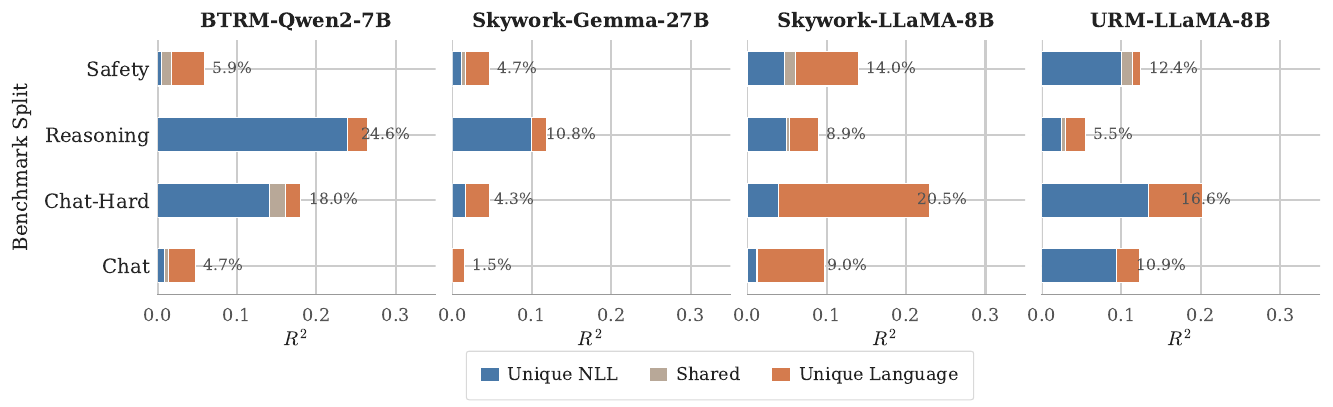}
  \caption{Stacked variance decomposition of reward model scores into three components: variance uniquely explained by NLL (blue), variance uniquely explained by language identity (orange), and variance shared between the two (gray). Percentages indicate total $R^2$ (sum of all three components). }
  \label{fig:variance_decomposition}
  
\end{figure}

Comparing M3 to M1 directly tests whether language identity retains predictive power after controlling for instance-level NLL. We ran incremental F-tests: Table~\ref{tab:structural_full} illustrates that adding language identity on top of NLL yielded significant improvements for every model evaluated. In addition, we present the stacked variance decomposition ($R^2$) in Figure \ref{fig:variance_decomposition}, that is, of all the variation in scores, how much can be explained by each factor.

Language identity remains a noticeable predictor independent of uncertainty, though its impact varies across models and tasks. For BTRM-Qwen2, NLL accounts for the largest share of explained variance, whereas for Skywork-LLaMA variance is mostly explained by language identity. It is also interesting to observe that while for BTRM-Qwen2 and Skywork-Gemma reasoning benchmarks' variances are mostly explained by NLL, variances from Chat and Safety are mostly explained by language identity.

\subsection{Can the Bias be Explained by Content Difficulty Alone?}
\label{sec:within_language}

\textbf{No: within-language, the content difficulty effect is inconsistent and domain-dependent.} If uncertainty were the only driving mechanism, then within any given language harder content (higher NLL) should consistently receive higher scores, and the bias would be reduced to a purely content-level effect.

\begin{figure}[t]
  \centering
  \includegraphics[width=0.6\columnwidth]{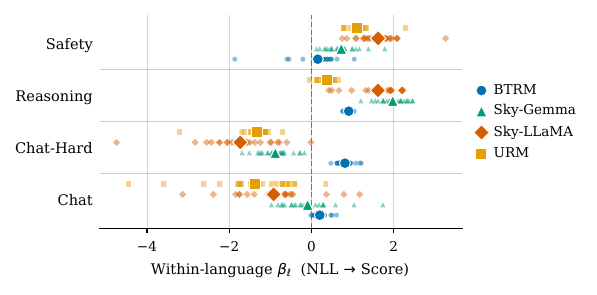}
  \caption{Distribution of within-language NLL$\to$Score slopes ($\beta_\ell$) across 4 models $\times$ 4 domains. Each dot is one model--domain pair; the x-axis shows the mean standardized slope.}
  \label{fig:within_language}
\end{figure}

To test this, we fit a separate regression within each of the 23 languages:
\begin{equation}
  \mathrm{Score}_{i\ell} = \alpha_\ell + \beta_\ell\,\mathrm{NLL}_{i\ell} + \varepsilon_{i\ell},
\end{equation}
and examine the distribution of per-language slopes $\beta_\ell$ across models and domains. If uncertainty were the driver, slopes should be consistently positive. Figure~\ref{fig:within_language} shows that they are not.

The relationship between within-language uncertainty and score is strongly domain-dependent. In Chat and Chat-Hard, slopes are in fact predominantly negative: higher NLL instances tend to receive \emph{lower} scores within the same language. Reward models also differ. Unlike BTRM which produces tightly clustered and consistent slopes in every domain, Sky-LLaMA and URM exhibit far more varied slopes. The full per-model breakdown is in Appendix Table~\ref{tab:within_language}.

\section{Conclusion and Future Directions}

This work demonstrates that multilingual evaluators exhibit large, systematic, and cross-model-consistent language-dependent shifts in pointwise scores when judging semantically matched content across 23 languages. Consequently, evaluators can appear robust under pairwise metrics while inducing substantial cross-language disparities in threshold-based decisions, particularly for lower-resource languages. As LLM evaluators are increasingly deployed for safety auditing and reward shaping, uncalibrated cross-lingual biases put the deployment of lower-resource languages at risk. Investigating this mechanism, we find that the bias correlates with model uncertainty but is not reducible to it. Language identity remains predictive after controlling for uncertainty, and within-language regressions show no consistent relationship between instance difficulty and scores, confirming that the bias operates at a structural, language level.

Practically, our results caution against treating raw evaluator scores as comparable across languages without calibration. A per-language offset correction, equivalent to language-specific thresholds, cuts the cross-language acceptance-rate gap by 60.9\% on average (Appendix~\ref{sec:appendix_score_priors}) and is the immediate takeaway for deployment, where global thresholds remain the default~\citep{NEURIPS2022_b1efde53, 51952}. It is not a complete fix. The residual gap has structure that no scalar offset reaches: score dispersion differs across languages, and 18.8\% of centered-score variance lies in language$\times$item interaction. Thresholding by language also requires language identification, which fails on code-switched inputs and opens an attack surface; wrapping Hindi content in an English frame raises acceptance from 50\% to 75\% under the mis-applied English threshold (\S\ref{sec:pairwise_bias}). More principled solutions therefore lie in training rather than post-hoc correction: language-aware calibration of reward models with multilingual data, normalization of language-conditioned scores during RLHF, language-balanced preference data during reward model training, and evaluation protocols that explicitly test cross-lingual score consistency alongside pairwise accuracy. We hope this work encourages the community to be aware of the multilingual bias in LLM evaluators, and for its validation, not only focus on pairwise accuracy but also take cross-language score consistency into account.

\subsubsection*{Acknowledgments}
This work was supported by the UKRI Frontier Grant EP/Y031350/1 (EQUATE).

\bibliography{colm2026_conference}
\bibliographystyle{colm2026_conference}

\clearpage

\appendix

\section{Experimental Details \& Dataset}
\label{sec:appendix}

Table~\ref{tab:language_info} lists the 23 languages included in our evaluation along with their scripts, language families, and resource classes.

The four \textsc{RewardBench} categories draw on over 20 independently constructed subsets: open-ended dialogue (AlpacaEval~\citealp{alpaca_eval}, MT-Bench~\citealp{NEURIPS2023_91f18a12}), adversarial robustness challenges (LLMBar~\citealp{zeng2024llmbar}), safety refusal and borderline-trigger prompts (XSTest~\citealp{rottger-etal-2024-xstest}, Do-Not-Answer~\citealp{wang-etal-2024-answer}), mathematical reasoning (PRM~\citealp{lightman2023letsverifystepstep}), and multilingual code correctness across six programming languages (HumanEvalPack~\citealp{muennighoff2023octopack}).

\begin{table}[h]
\centering
\small

\begin{tabular}{ccccc}
\toprule
\textbf{Code} & \textbf{Language} & \textbf{Script} & \textbf{Family} & \textbf{Resource Class} \\
\midrule
ar & Arabic      & Arabic        & Afro-Asiatic   & High \\
cs & Czech       & Latin         & Indo-European & High \\
de & German      & Latin         & Indo-European & High \\
el & Greek       & Greek         & Indo-European & Mid  \\
en & English     & Latin         & Indo-European & High \\
es & Spanish     & Latin         & Indo-European & High \\
fa & Persian     & Arabic        & Indo-European & High \\
fr & French      & Latin         & Indo-European & High \\
he & Hebrew      & Hebrew        & Afro-Asiatic  & Mid  \\
hi & Hindi       & Devanagari    & Indo-European & High \\
id & Indonesian  & Latin         & Austronesian  & Mid  \\
it & Italian     & Latin         & Indo-European & High \\
ja & Japanese    & Japanese      & Japonic       & High \\
ko & Korean      & Hangul        & Koreanic      & Mid  \\
nl & Dutch       & Latin         & Indo-European & High \\
pl & Polish      & Latin         & Indo-European & High \\
pt & Portuguese  & Latin         & Indo-European & High \\
ro & Romanian    & Latin         & Indo-European & Mid  \\
ru & Russian     & Cyrillic      & Indo-European & High \\
tr & Turkish     & Latin         & Turkic        & High \\
uk & Ukrainian   & Cyrillic      & Indo-European & Mid  \\
vi & Vietnamese  & Latin         & Austroasiatic & High \\
zh & Chinese     & Han / Hant    & Sino-Tibetan  & High \\
\bottomrule
\end{tabular}

\caption{Languages in RewardBench, M-RewardBench and their linguistic properties.
Data from \citet{gureja-etal-2025-rewardbench}.}
\label{tab:language_info}
\end{table}

\begin{figure}[h]
    \centering
    \includegraphics[width=0.63\linewidth]{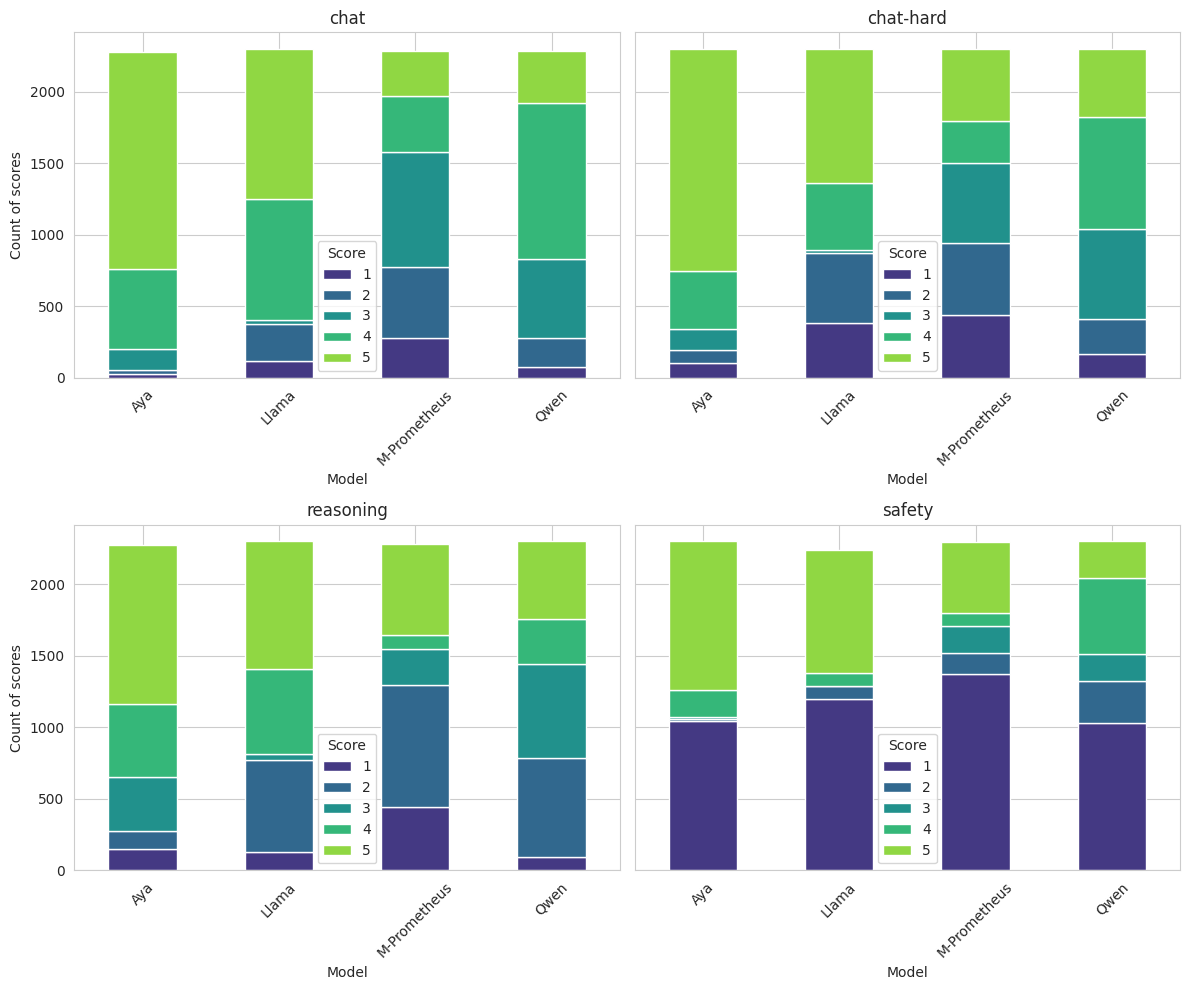}
    \caption{Distribution of reward scores for \textit{unchosen} responses across benchmark domains.
Stacked bars show the count of instances assigned to each discrete score level by different reward models.
Unchosen responses exhibit wider score dispersion than chosen responses, reducing ceiling effects and
revealing language-dependent variation.}
    \label{fig:score_distribution}
\end{figure}

For our experiment, we take \textit{prompt} and \textit{unchosen} as instruction-response pairs for our pointwise score analyses. As shown in Figure~\ref{fig:score_distribution}, the responses exhibit broad score distributions, and avoid the ceiling effects that would obscure language-dependent variation in evaluator behaviour.

\section{Supplementary Results for \S\ref{sec:experiments}}
\label{sec:appendix_scores}

This appendix collects per-language statistics, significance tests, cross-model correlations, and pairwise accuracy tables that support the main results in \S\ref{sec:lang_bias}--\ref{sec:pairwise_bias}.

\subsection{Per-Language Score Summaries}

Table~\ref{tab:variation_laaj} reports absolute (1--5) mean scores per language for prompted \textsc{LLM-as-a-Judge} models, broken down by benchmark split; Table~\ref{tab:variation_reward} reports the analogous per-language means for reward models after within-model z-normalization. Both tables include task-level and overall averages used to compute the language rankings discussed in \S\ref{sec:lang_bias}.

\begin{table*}[h]
\centering

\resizebox{1.0\textwidth}{!}{%
\begin{tabular}{
l
*{16}{S[table-format=1.2]}
*{5}{S[table-format=1.2]}
}
\toprule

& \multicolumn{16}{c}{\textbf{Model Scores}} 
& \multicolumn{5}{c}{\textbf{Task Averages}} \\
\cmidrule(lr){2-17} \cmidrule(lr){18-22}

\textbf{Lang}
& \multicolumn{4}{c}{Chat}
& \multicolumn{4}{c}{Chat-Hard}
& \multicolumn{4}{c}{Reasoning}
& \multicolumn{4}{c}{Safety}
& C & H & R & S & O \\
\cmidrule(lr){2-5}
\cmidrule(lr){6-9}
\cmidrule(lr){10-13}
\cmidrule(lr){14-17}
\cmidrule(lr){18-22}

& Aya & Q & M & LLaMA
& Aya & Q & M & LLaMA
& Aya & Q & M & LLaMA
& Aya & Q & M & LLaMA
& Avg & Avg & Avg & Avg & Avg \\
\midrule

hi & 4.64 & 4.01 & 3.75 & 4.15 & 4.74 & 3.98 & 3.56 & 4.04 & 4.50 & 3.55 & 2.83 & 3.93 & 3.41 & 2.72 & 2.57 & 2.94 & 4.14 & 4.08 & 3.70 & 2.91 & \textbf{3.71} \\
\rowcolor{gray!10}
el & 4.62 & 3.75 & 3.45 & 4.20 & 4.49 & 3.62 & 3.44 & 3.72 & 4.14 & 3.15 & 2.86 & 3.65 & 3.07 & 2.48 & 2.64 & 2.95 & 4.00 & 3.82 & 3.45 & 2.79 & 3.51 \\
fa & 4.45 & 3.72 & 3.10 & 4.06 & 4.46 & 3.76 & 3.21 & 3.88 & 4.25 & 3.27 & 2.95 & 3.65 & 3.20 & 2.61 & 2.52 & 2.88 & 3.83 & 3.83 & 3.53 & 2.80 & 3.50 \\
\rowcolor{gray!10}
vi & 4.71 & 3.78 & 3.04 & 4.08 & 4.46 & 3.68 & 3.29 & 3.48 & 4.05 & 3.29 & 2.90 & 3.63 & 3.27 & 2.54 & 2.29 & 2.96 & 3.90 & 3.73 & 3.47 & 2.76 & 3.46 \\
ko & 4.37 & 3.73 & 2.95 & 4.32 & 4.27 & 3.68 & 2.84 & 4.15 & 4.05 & 3.37 & 2.91 & 3.59 & 3.42 & 2.63 & 2.07 & 2.95 & 3.84 & 3.73 & 3.48 & 2.77 & 3.46 \\
\rowcolor{gray!10}
ja & 4.47 & 3.77 & 3.02 & 4.36 & 4.20 & 3.75 & 2.86 & 3.79 & 4.19 & 3.42 & 3.02 & 3.64 & 3.12 & 2.60 & 2.30 & 2.77 & 3.90 & 3.65 & 3.57 & 2.70 & 3.46 \\
he & 4.66 & 3.70 & 3.13 & 4.50 & 4.69 & 3.71 & 3.22 & 4.08 & 3.40 & 3.25 & 2.71 & 3.89 & 3.25 & 2.29 & 1.76 & 2.88 & 4.00 & 3.92 & 3.31 & 2.54 & 3.44 \\
\rowcolor{gray!10}
tr & 4.46 & 3.76 & 2.86 & 4.11 & 4.42 & 3.68 & 3.06 & 3.72 & 3.98 & 3.33 & 2.72 & 3.64 & 3.10 & 2.53 & 2.29 & 2.86 & 3.80 & 3.72 & 3.42 & 2.69 & 3.41 \\
ar & 4.60 & 3.78 & 2.95 & 4.11 & 4.58 & 3.73 & 2.88 & 3.72 & 4.03 & 3.32 & 2.78 & 3.73 & 3.00 & 2.49 & 1.67 & 2.84 & 3.86 & 3.73 & 3.46 & 2.50 & 3.39 \\
\rowcolor{gray!10}
uk & 4.74 & 3.64 & 3.02 & 4.05 & 4.58 & 3.62 & 3.20 & 3.60 & 4.02 & 3.20 & 2.77 & 3.75 & 2.94 & 2.39 & 2.19 & 2.20 & 3.86 & 3.75 & 3.44 & 2.43 & 3.37 \\
ru & 4.69 & 3.60 & 2.99 & 3.96 & 4.62 & 3.34 & 2.99 & 3.40 & 4.09 & 3.23 & 2.86 & 3.59 & 2.97 & 2.39 & 2.25 & 2.57 & 3.81 & 3.59 & 3.44 & 2.55 & 3.35 \\
\rowcolor{gray!10}
zh & 4.61 & 3.60 & 2.96 & 4.02 & 4.43 & 3.40 & 2.66 & 3.43 & 4.07 & 3.16 & 2.98 & 3.68 & 3.16 & 2.40 & 2.08 & 2.72 & 3.80 & 3.48 & 3.47 & 2.59 & 3.33 \\
ro & 4.51 & 3.64 & 3.05 & 4.06 & 4.31 & 3.41 & 3.04 & 3.36 & 3.63 & 3.17 & 2.93 & 3.65 & 3.00 & 2.34 & 2.45 & 2.60 & 3.82 & 3.53 & 3.35 & 2.60 & 3.32 \\
\rowcolor{gray!10}
id & 4.47 & 3.56 & 2.86 & 4.01 & 4.36 & 3.46 & 2.71 & 3.24 & 4.19 & 3.25 & 2.76 & 3.69 & 3.20 & 2.37 & 1.79 & 2.89 & 3.73 & 3.44 & 3.47 & 2.56 & 3.30 \\
nl & 4.48 & 3.50 & 3.04 & 3.99 & 4.24 & 3.27 & 3.11 & 3.11 & 3.99 & 3.13 & 2.90 & 3.48 & 3.12 & 2.36 & 2.24 & 2.50 & 3.75 & 3.43 & 3.38 & 2.56 & 3.28 \\
\rowcolor{gray!10}
pt & 4.54 & 3.57 & 2.73 & 4.04 & 4.27 & 3.32 & 2.77 & 3.25 & 3.94 & 3.16 & 2.68 & 3.56 & 3.01 & 2.38 & 2.19 & 2.69 & 3.72 & 3.40 & 3.33 & 2.57 & 3.26 \\
de & 4.60 & 3.57 & 2.94 & 3.89 & 4.32 & 3.23 & 2.70 & 3.10 & 4.18 & 3.15 & 2.89 & 3.48 & 2.98 & 2.33 & 2.26 & 2.46 & 3.75 & 3.34 & 3.42 & 2.51 & 3.25 \\
\rowcolor{gray!10}
en & 4.42 & 3.40 & 3.18 & 3.87 & 4.16 & 3.14 & 3.01 & 3.01 & 3.83 & 3.08 & 3.04 & 3.53 & 2.94 & 2.26 & 2.43 & 2.72 & 3.72 & 3.33 & 3.37 & 2.59 & 3.25 \\
cs & 4.54 & 3.62 & 2.76 & 3.95 & 4.28 & 3.46 & 2.66 & 3.44 & 4.25 & 3.11 & 2.57 & 3.61 & 2.56 & 2.45 & 2.12 & 2.61 & 3.72 & 3.46 & 3.38 & 2.44 & 3.25 \\
\rowcolor{gray!10}
es & 4.38 & 3.53 & 2.76 & 3.98 & 4.24 & 3.34 & 2.75 & 3.08 & 3.80 & 3.18 & 2.85 & 3.68 & 3.04 & 2.33 & 2.38 & 2.56 & 3.66 & 3.35 & 3.38 & 2.58 & 3.24 \\
pl & 4.57 & 3.66 & 2.89 & 3.89 & 4.38 & 3.47 & 2.95 & 3.19 & 3.61 & 3.12 & 2.90 & 3.59 & 2.88 & 2.36 & 2.06 & 2.35 & 3.75 & 3.50 & 3.31 & 2.41 & 3.24 \\
\rowcolor{gray!10}
fr & 4.43 & 3.51 & 2.70 & 4.00 & 4.22 & 3.25 & 2.69 & 3.06 & 3.88 & 3.14 & 2.75 & 3.59 & 3.07 & 2.38 & 2.37 & 2.65 & 3.66 & 3.31 & 3.34 & 2.62 & 3.23 \\
it & 4.53 & 3.41 & 2.59 & 3.92 & 4.40 & 3.28 & 2.58 & 3.01 & 4.08 & 3.18 & 2.78 & 3.50 & 2.97 & 2.32 & 2.06 & 2.56 & 3.61 & 3.32 & 3.38 & 2.48 & \textbf{3.20} \\
\bottomrule
\end{tabular}

}
\caption{Average evaluation scores (\textbf{1-5}) across languages for four task categories (Chat, Chat-Hard, Reasoning, Safety) and four models (Aya, Qwen, Unbabel, LLaMA). Task-level and overall averages are reported for each language.}
\label{tab:variation_laaj}
\end{table*}

\begin{table*}[h]
\centering
\resizebox{1.0\textwidth}{!}{%
\begin{tabular}{
l
*{16}{S[table-format=-1.2]}
*{5}{S[table-format=-1.2]}
}
\toprule

& \multicolumn{16}{c}{\textbf{Model Scores}}
& \multicolumn{5}{c}{\textbf{Task Averages}} \\
\cmidrule(lr){2-17} \cmidrule(lr){18-22}

\textbf{Lang} & \multicolumn{4}{c}{Chat} & \multicolumn{4}{c}{Chat-Hard} & \multicolumn{4}{c}{Reasoning} & \multicolumn{4}{c}{Safety} & C & H & R & S & O \\
\cmidrule(lr){2-5}\cmidrule(lr){6-9}\cmidrule(lr){10-13}\cmidrule(lr){14-17}\cmidrule(lr){18-22}

 & URM & BTRM & SkyG & SkyL & URM & BTRM & SkyG & SkyL & URM & BTRM & SkyG & SkyL & URM & BTRM & SkyG & SkyL & Avg & Avg & Avg & Avg & Avg \\
\midrule

el & 1.58 & 1.14 & 0.62 & 1.59 & 0.46 & 0.97 & 0.92 & 0.84 & 1.78 & -0.11 & 0.35 & 1.22 & -0.28 & -0.70 & -1.14 & -0.62 & 1.23 & 0.80 & 0.81 & -0.69 & \textbf{0.54} \\
\rowcolor{gray!10}
he & 1.56 & 0.66 & 0.67 & 1.14 & 0.10 & 0.63 & 0.66 & 0.58 & 1.79 & -0.11 & 0.36 & 0.75 & 0.26 & -1.50 & -1.59 & -1.24 & 1.01 & 0.49 & 0.70 & -1.02 & 0.29 \\
ar & 0.95 & 1.21 & 0.61 & 1.09 & -0.35 & 0.95 & 0.50 & 0.45 & 1.39 & 0.18 & 0.31 & 1.00 & -0.95 & -1.20 & -1.70 & -0.84 & 0.96 & 0.39 & 0.72 & -1.17 & 0.22 \\
\rowcolor{gray!10}
hi & 0.82 & 0.99 & 0.73 & 0.92 & -0.15 & 0.96 & 0.80 & 0.33 & 0.96 & 0.11 & 0.38 & 0.75 & -0.65 & -0.87 & -1.46 & -1.12 & 0.86 & 0.48 & 0.55 & -1.02 & 0.22 \\
uk & 0.42 & 0.97 & 0.86 & 1.36 & -0.39 & 0.84 & 0.75 & 0.77 & 1.27 & -0.16 & 0.38 & 1.17 & -1.23 & -1.43 & -1.43 & -0.75 & 0.90 & 0.49 & 0.66 & -1.21 & 0.21 \\
\rowcolor{gray!10}
ro & 1.01 & 0.86 & 0.63 & 1.08 & -0.39 & 0.79 & 0.71 & 0.45 & 1.37 & -0.08 & 0.35 & 1.09 & -1.04 & -1.40 & -1.66 & -1.05 & 0.89 & 0.39 & 0.68 & -1.29 & 0.17 \\
fa & 0.90 & 1.07 & 0.36 & 0.67 & -0.10 & 1.05 & 0.49 & 0.25 & 1.48 & 0.01 & 0.14 & 0.95 & -0.80 & -0.95 & -1.58 & -1.22 & 0.75 & 0.43 & 0.64 & -1.14 & 0.17 \\
\rowcolor{gray!10}
tr & 0.58 & 0.82 & 1.07 & 0.70 & -0.66 & 0.85 & 0.95 & 0.18 & 1.11 & -0.06 & 0.57 & 0.76 & -1.25 & -1.18 & -1.33 & -1.05 & 0.79 & 0.33 & 0.59 & -1.20 & 0.13 \\
cs & 0.42 & 1.04 & 0.90 & 1.06 & -0.65 & 0.89 & 0.66 & 0.25 & 1.03 & -0.06 & 0.43 & 0.90 & -0.97 & -1.46 & -1.47 & -0.95 & 0.85 & 0.29 & 0.57 & -1.21 & 0.13 \\
\rowcolor{gray!10}
pl & 0.08 & 1.03 & 0.74 & 1.04 & -0.55 & 0.79 & 0.42 & 0.29 & 1.10 & 0.04 & 0.65 & 0.98 & -1.07 & -1.29 & -1.46 & -1.10 & 0.72 & 0.24 & 0.69 & -1.23 & 0.11 \\
vi & 0.66 & 0.99 & 0.49 & 1.06 & -0.71 & 0.76 & 0.58 & 0.34 & 1.27 & 0.03 & 0.32 & 0.88 & -1.19 & -1.44 & -1.73 & -1.09 & 0.80 & 0.24 & 0.62 & -1.36 & 0.08 \\
\rowcolor{gray!10}
ko & 0.84 & 0.75 & 1.15 & 0.30 & -0.55 & 0.67 & 1.14 & -0.47 & 1.27 & -0.07 & 0.78 & 0.39 & -1.01 & -1.41 & -0.86 & -1.73 & 0.76 & 0.20 & 0.59 & -1.25 & 0.07 \\
ru & 0.72 & 0.98 & 0.60 & 1.08 & -0.60 & 0.67 & 0.45 & 0.19 & 1.05 & 0.04 & 0.59 & 0.85 & -1.22 & -1.48 & -1.58 & -1.22 & 0.85 & 0.18 & 0.63 & -1.37 & 0.07 \\
\rowcolor{gray!10}
nl & 0.70 & 0.90 & 0.55 & 0.77 & -0.62 & 0.69 & 0.09 & -0.19 & 1.04 & -0.01 & 0.68 & 0.77 & -1.03 & -1.52 & -1.94 & -1.45 & 0.73 & -0.01 & 0.62 & -1.49 & -0.04 \\
it & 0.38 & 1.09 & 0.47 & 0.97 & -1.21 & 0.67 & 0.31 & -0.01 & 1.01 & 0.12 & 0.49 & 0.83 & -1.46 & -1.44 & -1.96 & -1.30 & 0.72 & -0.06 & 0.61 & -1.54 & -0.07 \\
\rowcolor{gray!10}
de & 0.54 & 0.87 & 0.68 & 0.55 & -1.04 & 0.65 & 0.38 & -0.43 & 0.86 & -0.00 & 0.57 & 0.53 & -1.14 & -1.62 & -1.80 & -1.60 & 0.66 & -0.11 & 0.49 & -1.54 & -0.13 \\
ja & 0.08 & 0.66 & 0.72 & 0.42 & -0.86 & 0.60 & 0.63 & -0.26 & 1.13 & -0.08 & 0.31 & 0.60 & -1.54 & -1.90 & -1.12 & -1.70 & 0.47 & 0.03 & 0.49 & -1.57 & -0.14 \\
\rowcolor{gray!10}
pt & 0.56 & 0.89 & 0.45 & 0.30 & -0.94 & 0.55 & 0.28 & -0.59 & 1.09 & 0.10 & 0.65 & 0.46 & -1.18 & -1.62 & -2.02 & -1.85 & 0.55 & -0.17 & 0.57 & -1.67 & -0.18 \\
es & 0.11 & 0.96 & 0.57 & 0.54 & -1.44 & 0.61 & -0.02 & -0.53 & 0.82 & 0.12 & 0.45 & 0.52 & -1.54 & -1.57 & -2.23 & -1.65 & 0.54 & -0.35 & 0.48 & -1.75 & -0.27 \\
\rowcolor{gray!10}
fr & 0.18 & 0.88 & 0.49 & 0.71 & -1.52 & 0.51 & 0.25 & -0.53 & 0.48 & -0.02 & 0.64 & 0.55 & -1.83 & -1.69 & -2.09 & -1.74 & 0.57 & -0.32 & 0.41 & -1.84 & -0.30 \\
zh & 0.67 & -0.05 & 0.77 & 0.09 & -1.08 & -0.08 & 0.79 & -0.90 & 1.08 & -0.02 & 0.60 & 0.19 & -1.34 & -3.10 & -1.49 & -2.18 & 0.37 & -0.32 & 0.47 & -2.03 & -0.38 \\
\rowcolor{gray!10}
id & -0.06 & 0.61 & 0.34 & 0.15 & -0.80 & 0.42 & -0.13 & -0.67 & 0.88 & -0.02 & 0.24 & 0.32 & -1.25 & -2.10 & -2.06 & -1.93 & 0.26 & -0.30 & 0.35 & -1.84 & -0.38 \\
en & 0.71 & 0.58 & 0.75 & -0.21 & -1.49 & 0.18 & 0.13 & -2.09 & 0.67 & -0.12 & 0.83 & 0.05 & -1.10 & -2.47 & -2.29 & -2.72 & 0.46 & -0.82 & 0.36 & -2.15 & \textbf{-0.54} \\
\bottomrule
\end{tabular}
}
\caption{Average evaluation normalized z-scores across languages for four task categories (Chat, Chat-Hard, Reasoning, Safety) and four reward models (URM-LLaMa, BTRM-Qwen, Skywork-Gemma, Skywork-LLaMa). Scores are z-normalized within each model. Task-level and overall averages are reported for each language.}
\label{tab:variation_reward}
\end{table*}

\subsection{Statistical Significance of Language-Dependent Effects}
Table~\ref{tab:anova} reports one-way ANOVA tests of equal mean scores across languages for each evaluator (prompted judges and reward models), confirming that all language effects are statistically significant ($p<0.001$; \S\ref{sec:lang_bias}).

\begin{table}[h]
\centering
\small
\setlength{\tabcolsep}{5pt}
\renewcommand{\arraystretch}{1.15}

\begin{tabularx}{\linewidth}{@{}lccc|lccc@{}}
\toprule
\textbf{LLM-as-a-Judge} & $F$ & $p$-value & $n$ & \textbf{Reward Model} & $F$ & $p$-value & $n$ \\
\midrule
Aya-Expanse-32B           & 2.31  & $<0.001$ & 9148 &
RM-LLaMA-3.1-8B           & 7.25  & $<0.001$ & 9200 \\
Qwen-2.5-72B              & 4.45  & $<0.001$ & 9187 &
BTRM-Qwen-2-7B            & 8.42  & $<0.001$ & 9200 \\
M-Prometheus-14B          & 5.38  & $<0.001$ & 9161 &
Skywork-Gemma-2-27B       & 4.10  & $<0.001$ & 9200 \\
LLaMA-3.1-70B-Instruct    & 4.83  & $<0.001$ & 9138 &
Skywork-LLaMA-3.1-8B & 29.10 & $<0.001$ & 9200 \\
\bottomrule
\end{tabularx}

\caption{One-way ANOVA results testing for differences in mean evaluation scores across languages.
All evaluators exhibit statistically significant language effects ($p<0.001$).}
\label{tab:anova}
\end{table}

\subsection{Cross-Model Correlation Matrices}
\begin{figure*}[h]
    \centering
    \begin{subfigure}[t]{0.44\linewidth}
        \centering
    \includegraphics[width=\linewidth]{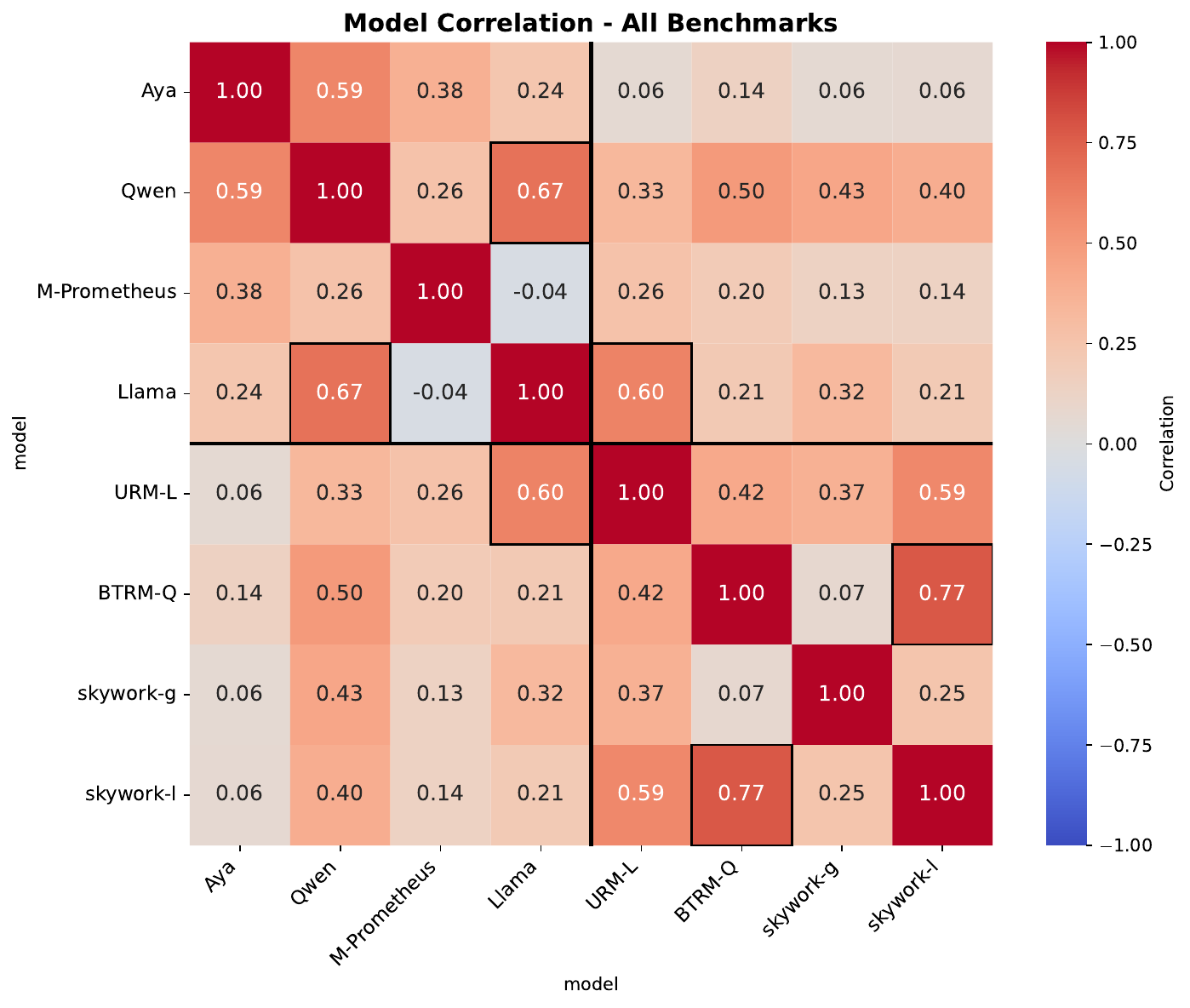}
    \end{subfigure}
    \hfill
    \begin{subfigure}[t]{0.54\linewidth}
        \centering
        \includegraphics[width=\linewidth]{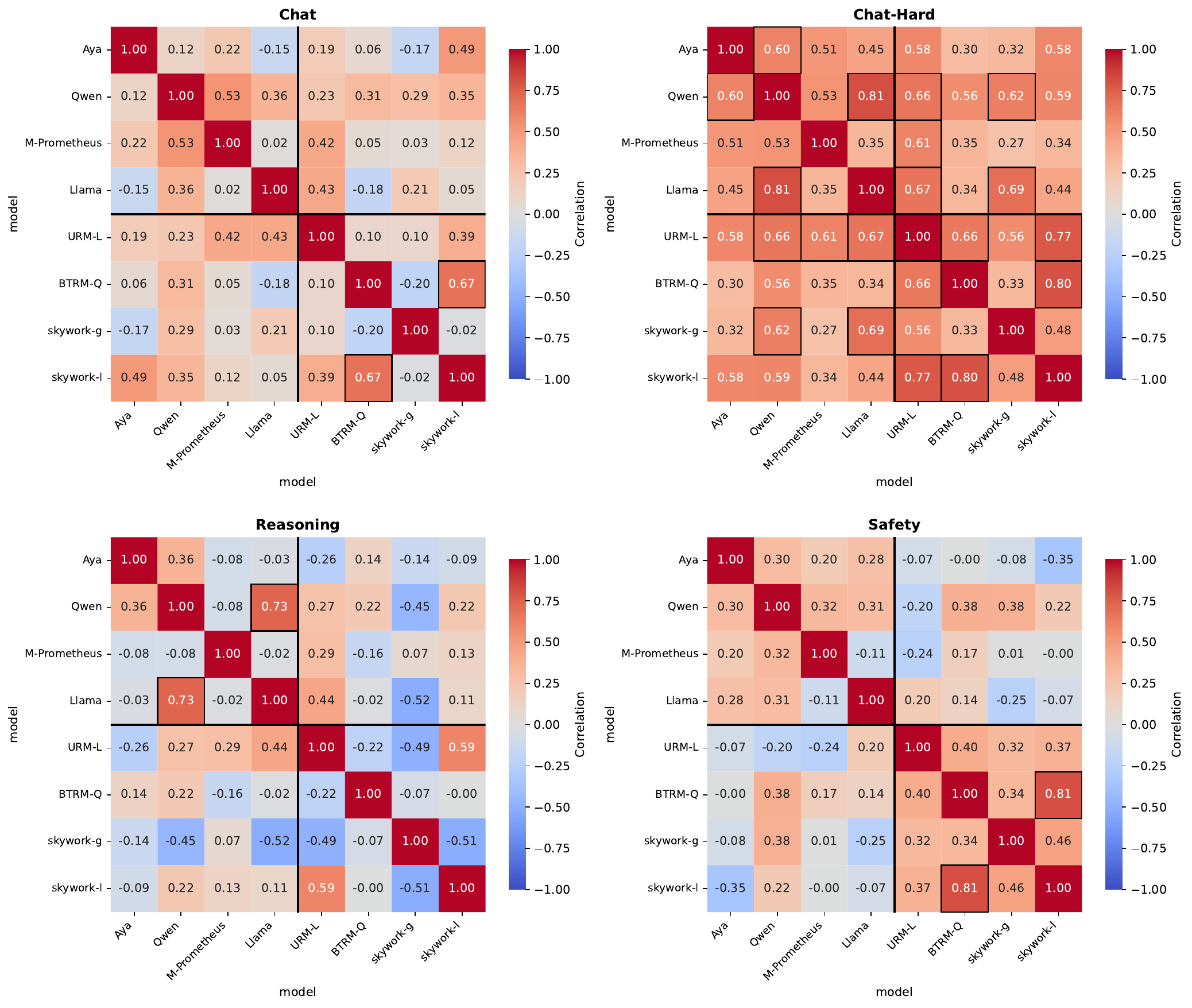}
    \end{subfigure}
    \caption{Correlation of language-dependent scoring patterns across judge models, aggregated over all benchmarks. Each cell reports the Pearson correlation between per-language mean scores assigned by two models.}
    \label{fig:model_correlation}
\end{figure*}

Figure~\ref{fig:model_correlation} visualizes pairwise correlations between evaluators' \emph{per-language mean scores}, supporting the cross-model consistency claim in \S\ref{cross_model_consistency}.

\subsection{Per-Model Pairwise Accuracy Tables}

Table~\ref{tab:pairwise_accuracy_avg} reports per-language pairwise accuracy for both prompted judges and reward models. This table supports the claim in \S\ref{sec:pairwise_bias} that pairwise accuracy remains high and stable even when pointwise scores exhibit large language-dependent shifts.

\begin{table}[h]
\centering
\small
\resizebox{!}{0.25\linewidth}{
\begin{tabular}{lrrrrrrrrrrrrrrrr}
\toprule

& \multicolumn{8}{c}{\textbf{LLM-as-a-Judge}} & \multicolumn{8}{c}{\textbf{Reward Models}} \\
\cmidrule(r){2-9} \cmidrule(r){10-17}
Language & \multicolumn{2}{c}{Aya} & \multicolumn{2}{c}{Qwen} & \multicolumn{2}{c}{M-Prom} & \multicolumn{2}{c}{Llama 3} & \multicolumn{2}{c}{BTRM-Q} & \multicolumn{2}{c}{URM-L} & \multicolumn{2}{c}{Skywork-g} & \multicolumn{2}{c}{Skywork-l} \\
 & Chat & Safety & Chat & Safety & Chat & Safety & Chat & Safety & Chat & Safety & Chat & Safety & Chat & Safety & Chat & Safety \\
\midrule

\textbf{English} & 81.4 & 83.3 & 93.0 & 87.0 & 91.6 & 77.6 & 87.6 & 83.9 & 98.0 & 92.0 & 97.0 & 90.0 & 88.0 & 93.0 & 99.0 & 93.0 \\
\rowcolor{gray!10}
Russian & 76.1 & 83.8 & 91.0 & 84.7 & 91.8 & 78.5 & 85.6 & 81.5 & 97.0 & 85.0 & 92.0 & 88.0 & 84.0 & 86.0 & 94.0 & 91.0 \\
Italian & 79.4 & 83.7 & 93.6 & 83.8 & 93.9 & 76.8 & 85.4 & 85.2 & 98.0 & 85.0 & 89.0 & 86.0 & 85.0 & 85.0 & 97.0 & 91.0 \\
\rowcolor{gray!10}
German & 78.4 & 82.7 & 92.0 & 85.0 & 90.8 & 74.7 & 86.1 & 82.8 & 98.0 & 89.0 & 92.0 & 85.0 & 83.0 & 85.0 & 93.0 & 91.0 \\
Portuguese & 79.1 & 86.0 & 91.4 & 83.9 & 93.0 & 75.6 & 85.5 & 83.0 & 98.0 & 86.0 & 91.0 & 86.0 & 85.0 & 85.0 & 93.0 & 89.0 \\
\rowcolor{gray!10}
Romanian & 78.2 & 84.4 & 90.1 & 83.6 & 92.2 & 74.7 & 85.1 & 83.2 & 96.0 & 86.0 & 89.0 & 81.0 & 86.0 & 84.0 & 91.0 & 86.0 \\
Indonesian & 80.0 & 80.3 & 92.3 & 86.0 & 92.7 & 77.0 & 87.4 & 80.8 & 98.0 & 84.0 & 87.0 & 84.0 & 78.0 & 88.0 & 89.0 & 92.0 \\
\rowcolor{gray!10}
Spanish & 82.6 & 81.0 & 92.3 & 84.2 & 93.0 & 72.7 & 84.9 & 83.9 & 98.0 & 88.0 & 90.0 & 85.0 & 78.0 & 87.0 & 93.0 & 90.0 \\
Vietnamese & 75.7 & 81.4 & 87.9 & 79.7 & 88.6 & 74.6 & 85.5 & 81.2 & 97.0 & 85.0 & 87.0 & 82.0 & 87.0 & 88.0 & 92.0 & 83.0 \\
\rowcolor{gray!10}
Chinese & 78.2 & 79.9 & 89.4 & 82.7 & 89.2 & 76.6 & 85.0 & 78.6 & 90.0 & 85.0 & 86.0 & 83.0 & 84.0 & 83.0 & 91.0 & 85.0 \\
Turkish & 79.9 & 77.8 & 92.2 & 86.7 & 92.2 & 76.9 & 85.9 & 81.2 & 97.0 & 82.0 & 86.0 & 80.0 & 82.0 & 84.0 & 87.0 & 82.0 \\
\rowcolor{gray!10}
French & 80.0 & 83.5 & 92.7 & 85.1 & 92.3 & 74.9 & 84.2 & 85.1 & 98.0 & 88.0 & 85.0 & 85.0 & 85.0 & 91.0 & 91.0 & 90.0 \\
Korean & 78.4 & 77.1 & 90.1 & 82.5 & 92.2 & 78.4 & 79.0 & 72.6 & 97.0 & 85.0 & 90.0 & 85.0 & 81.0 & 83.0 & 92.0 & 80.0 \\
\rowcolor{gray!10}
Dutch & 80.5 & 82.5 & 92.9 & 86.0 & 91.1 & 81.5 & 85.1 & 83.2 & 98.0 & 88.0 & 84.0 & 87.0 & 83.0 & 87.0 & 94.0 & 90.0 \\
Polish & 78.6 & 84.2 & 88.7 & 86.3 & 89.1 & 78.2 & 86.9 & 83.9 & 97.0 & 84.0 & 83.0 & 85.0 & 78.0 & 87.0 & 90.0 & 88.0 \\
\rowcolor{gray!10}
Arabic & 77.8 & 80.9 & 88.3 & 85.8 & 92.5 & 74.2 & 85.1 & 82.9 & 97.0 & 85.0 & 83.0 & 80.0 & 76.0 & 86.0 & 85.0 & 79.0 \\
Greek & 77.3 & 79.8 & 92.4 & 82.3 & 86.7 & 79.5 & 80.4 & 79.5 & 94.0 & 80.0 & 85.0 & 81.0 & 73.0 & 81.0 & 85.0 & 84.0 \\
\rowcolor{gray!10}
Czech & 75.9 & 81.8 & 91.4 & 80.9 & 89.9 & 77.5 & 86.9 & 83.0 & 98.0 & 87.0 & 82.0 & 82.0 & 80.0 & 83.0 & 90.0 & 85.0 \\
Ukrainian & 74.4 & 78.1 & 90.1 & 81.1 & 88.5 & 74.1 & 84.5 & 83.5 & 96.0 & 84.0 & 86.0 & 84.0 & 80.0 & 77.0 & 91.0 & 87.0 \\
\rowcolor{gray!10}
Hindi & 76.3 & 81.5 & 91.4 & 84.1 & 84.8 & 70.1 & 85.1 & 79.8 & 93.0 & 80.0 & 88.0 & 83.0 & 75.0 & 82.0 & 83.0 & 85.0 \\
Japanese & 77.7 & 82.2 & 91.8 & 84.7 & 91.6 & 72.2 & 79.3 & 80.2 & 95.0 & 87.0 & 86.0 & 85.0 & 70.0 & 81.0 & 90.0 & 89.0 \\
\rowcolor{gray!10}
Persian & 80.9 & 81.8 & 89.9 & 85.4 & 90.1 & 76.4 & 86.7 & 79.3 & 96.0 & 80.0 & 78.0 & 85.0 & 81.0 & 85.0 & 88.0 & 86.0 \\
Hebrew & 76.7 & 77.5 & 87.1 & 81.7 & 86.8 & 66.6 & 77.2 & 78.3 & 97.0 & 85.0 & 56.0 & 75.0 & 82.0 & 82.0 & 74.0 & 85.0 \\
\midrule
\rowcolor{gray!10}
Mean & \multicolumn{2}{c}{80.0} & \multicolumn{2}{c}{87.5} & \multicolumn{2}{c}{83.1} & \multicolumn{2}{c}{83.1} & \multicolumn{2}{c}{90.9} & \multicolumn{2}{c}{84.8} & \multicolumn{2}{c}{83.0} & \multicolumn{2}{c}{88.5} \\
Std Dev & \multicolumn{2}{c}{2.7} & \multicolumn{2}{c}{3.9} & \multicolumn{2}{c}{8.0} & \multicolumn{2}{c}{3.1} & \multicolumn{2}{c}{6.2} & \multicolumn{2}{c}{5.8} & \multicolumn{2}{c}{4.4} & \multicolumn{2}{c}{4.7} \\

\bottomrule
\end{tabular}
}
\caption{Per-language pairwise accuracy (accepted $\geq$ rejected) for the Chat and Safety benchmarks. Despite substantial language-dependent variation in pointwise model scores, pairwise accuracy remains uniformly high and consistent across languages and models, with minimal variation ($\leq 5\%$) within each evaluator.}
\label{tab:pairwise_accuracy_avg}
\end{table}

\section{Supplementary Analyses for \S\ref{sec:mechanistic}}
\label{sec:appendix_analyses}

\subsection{Uncertainty--Score Correlations by Model and Split}
\label{sec:appendix_uncertainty_corr}

Table~\ref{tab:nll_score_correlation} reports language-level Spearman and Pearson correlations between mean NLL and mean reward score, computed separately for each (model, split) combination. This table provides the per-setting breakdown referenced in \S\ref{sec:nll_score}.

\begin{table}[h]
\centering
\small

\begin{tabular}{llcc}
\toprule
\textbf{Category} & \textbf{Model} & \textbf{Spearman $\rho$} & \textbf{Pearson $r$} \\
\midrule
  chat & BTRM-Qwen2-7B & 0.582$^{**}$ & 0.519 \\
  chat & Skywork-Gemma-27B & 0.172 & 0.100 \\
  chat & Skywork-LLaMA-8B & 0.581$^{**}$ & 0.529 \\
  chat & URM-LLaMA-3.1-8B & 0.120 & 0.522 \\
  chat-hard & BTRM-Qwen2-7B & 0.862$^{***}$ & 0.720 \\
  chat-hard & Skywork-Gemma-27B & 0.633$^{**}$ & 0.585 \\
  chat-hard & Skywork-LLaMA-8B & 0.554$^{**}$ & 0.545 \\
  chat-hard & URM-LLaMA-3.1-8B & 0.499$^{*}$ & 0.626 \\
  reasoning & BTRM-Qwen2-7B & -0.140 & -0.101 \\
  reasoning & Skywork-Gemma-27B & -0.660$^{***}$ & -0.705 \\
  reasoning & Skywork-LLaMA-8B & 0.456$^{*}$ & 0.324 \\
  reasoning & URM-LLaMA-3.1-8B & 0.233 & 0.517 \\
  safety & BTRM-Qwen2-7B & 0.812$^{***}$ & 0.714 \\
  safety & Skywork-Gemma-27B & 0.646$^{***}$ & 0.662 \\
  safety & Skywork-LLaMA-8B & 0.356 & 0.404 \\
  safety & URM-LLaMA-3.1-8B & 0.387 & 0.772 \\
\midrule
\multicolumn{2}{l}{\textit{Mean across all runs}} & 0.381 & 0.421 \\
\bottomrule
\end{tabular}
\caption{Language-level Spearman correlation between mean total NLL and mean reward score across all model--dataset combinations. Positive $\rho$ indicates that languages with higher NLL (greater model uncertainty) receive higher reward scores. Significance: $^{***}p<0.001$, $^{**}p<0.01$, $^{*}p<0.05$.}
\label{tab:nll_score_correlation}
\end{table}

\subsection{Summed versus Per-Token NLL}
\label{sec:appendix_nll_metric}

Per-token NLL divides total surprisal by token count, and both quantities grow together for lower-resource languages: the model is more surprised by the content, and the tokenizer segments the same content into more pieces~\citep{rust-etal-2021-good, ahia-etal-2023-languages, petrov2023token_unfairness}. The ratio therefore suppresses the uncertainty signal. Table~\ref{tab:summed_vs_pertoken} illustrates this under URM-LLaMA-3.1-8B: summed NLL separates Hebrew (most uncertain of the 23 languages) from English (most confident) by a factor of 2.6, while per-token NLL is essentially flat and does not preserve the ordering.

\begin{table}[h]
\centering
\small
\begin{tabular}{lccc}
\toprule
\textbf{Language} & \textbf{Mean summed NLL} & \textbf{Rank (summed)} & \textbf{Mean per-token NLL} \\
\midrule
Hebrew  & 9{,}441 & 1  & 33.97 \\
Hindi   & 7{,}018 & 2  & 33.45 \\
Italian & 4{,}873 & 11 & 32.45 \\
Arabic  & 4{,}853 & 12 & 31.55 \\
English & 3{,}603 & 23 & 33.52 \\
\bottomrule
\end{tabular}
\caption{Summed versus per-token NLL under URM-LLaMA-3.1-8B (LLaMA-3.1-8B base backbone), aggregated across the four benchmark splits. Rank 1 = highest summed NLL among the 23 languages.}
\label{tab:summed_vs_pertoken}
\end{table}

It is well established that multilingual LLMs perform unevenly across languages~\citep{ahuja-etal-2023-mega}, and language-conditioned calibration failures have been observed directly~\citep{zhou2025finallayerintermediaterepresentations}. Summed NLL avoids the token-count denominator entirely. It is the model-assigned $-\log P$ of the full string, a property of the (model, text) pair that is invariant to how the string is segmented; since \textsc{M-RewardBench} is parallel by design, comparing it across languages compares the model's surprise at the same semantic content.

Table~\ref{tab:summed_nll_by_model} reports the per-language summed-NLL pattern under each backbone, with the language-level Spearman correlation between mean summed NLL and mean reward score. Lower-resource, non-Latin-script languages (Hebrew, Hindi, Greek, Farsi) are consistently the most uncertain and English the most confident. The correlation is positive for all four models and significant for three; URM's summed-NLL correlation is the weakest ($\rho = +0.35$, $p = 0.10$), while its internal, token-free uncertainty signals show a stronger association (Appendix~\ref{sec:appendix_token_free}).

\begin{table}[h]
\centering
\small
\resizebox{\columnwidth}{!}{%
\begin{tabular}{lcccc}
\toprule
\textbf{Backbone} & \textbf{Summed NLL range} & \textbf{Most uncertain} & \textbf{Most confident} & \textbf{$\rho$(NLL, score)} \\
\midrule
BTRM-Qwen-2-7B     & 4{,}850--15{,}010 & Greek, Hindi  & English, Chinese & $+0.67$ ($p<0.001$) \\
URM-LLaMA-3.1-8B   & 3{,}603--9{,}441  & Hebrew, Hindi & English, Chinese & $+0.35$ ($p=0.10$)  \\
Skywork-Gemma-27B  & 559--954          & Farsi, Greek  & English, Spanish & $+0.51$ ($p=0.014$) \\
Skywork-LLaMA-8B   & 3{,}616--9{,}491  & Hebrew, Hindi & English, Chinese & $+0.52$ ($p=0.010$) \\
\bottomrule
\end{tabular}%
}
\caption{Per-language mean summed NLL under each base backbone and its language-level Spearman correlation with mean reward score ($n=23$ languages).}
\label{tab:summed_nll_by_model}
\end{table}

\subsection{Alternative Uncertainty Measures}
\label{sec:appendix_token_free}

Table~\ref{tab:token_free} (main text) reports four uncertainty measures computed without token probabilities; we detail their operationalization here.

\paragraph{URM internal signals.} URM-LLaMA-3.1-8B predicts five reward attributes through a Gaussian value head~\citep{lou2025uncertaintyawarerewardmodelteaching}, so its uncertainty can be read off internal signals directly. \emph{Attribute-head disagreement} is the weighted standard deviation across the five attribute means, treating the attributes as a deep ensemble~\citep{lakshminarayanan2017ensembles}; the unweighted standard deviation and the max--min range give similar language-level correlations ($\rho = +0.54$ and $+0.69$). \emph{Predictive variance} is the combined Gaussian $\sigma$ across the five attribute heads~\citep{gal2016dropout}.

\paragraph{Semantic entropy.} For the prompted judges (AWQ-quantized Qwen-2.5-72B-Instruct and LLaMA-3.1-70B-Instruct), we follow \citet{kuhn2023semanticuncertainty}: sample $M=5$ chain-of-thought judgments per instance, cluster the samples by NLI-based semantic equivalence, and take the Shannon entropy over cluster assignments.

\subsection{Full Structural Decomposition Details}
\label{sec:appendix_decomposition}

Table~\ref{tab:structural_full} reports the full nested-model variance decomposition (NLL-only, language-only, and full models) and incremental $F$-tests, split by model and benchmark category.

\begin{table*}[h]
\centering

\small
\resizebox{\textwidth}{!}{%
\begin{tabular}{ll ccc cc cc}
\toprule
 & & \multicolumn{3}{c}{\textbf{$R^2$}} & \multicolumn{2}{c}{\textbf{Lang\,$|$\,NLL}} & \multicolumn{2}{c}{\textbf{NLL\,$|$\,Lang}} \\
\cmidrule(lr){3-5} \cmidrule(lr){6-7} \cmidrule(lr){8-9}
\textbf{Model} & \textbf{Split} & NLL & Lang & Full & $\Delta R^2$ & $F$ & $\Delta R^2$ & $F$ \\
\midrule
\multirow{4}{*}{BTRM-Qwen2-7B}
 & Chat      & .014 & .039 & .047 & .033 & 3.6***  & .008 & 20.0***  \\
 & Chat-Hard & .161 & .040 & .180 & .019 & 2.4***  & .141 & 390.8*** \\
 & Reasoning & .219 & .007 & .246 & .026 & 3.6***  & .239 & 719.8*** \\
 & Safety    & .017 & .055 & .059 & .042 & 4.6***  & .005 & 11.0***  \\
\midrule
\multirow{4}{*}{URM-LLaMA-3.1-8B}
 & Chat      & .080 & .015 & .109 & .029 & 3.4***  & .094 & 240.1*** \\
 & Chat-Hard & .097 & .032 & .166 & .069 & 8.5***  & .134 & 365.0*** \\
 & Reasoning & .030 & .030 & .055 & .025 & 2.7***  & .025 & 60.2***  \\
 & Safety    & .115 & .024 & .124 & .010 & 1.2     & .100 & 260.1*** \\
\midrule
\multirow{4}{*}{Skywork-Gemma-27B}
 & Chat      & .000 & .015 & .015 & .015 & 1.6*    & .000 & 0.5      \\
 & Chat-Hard & .013 & .026 & .043 & .030 & 3.2***  & .017 & 40.2***  \\
 & Reasoning & .089 & .009 & .108 & .019 & 2.2***  & .100 & 254.5*** \\
 & Safety    & .017 & .035 & .047 & .030 & 3.3***  & .012 & 27.9***  \\
\midrule
\multirow{4}{*}{Skywork-LLaMA-8B}
 & Chat      & .005 & .078 & .090 & .085 & 9.7***  & .012 & 31.1***  \\
 & Chat-Hard & .014 & .165 & .205 & .191 & 24.8*** & .039 & 111.9*** \\
 & Reasoning & .053 & .040 & .089 & .036 & 4.1***  & .049 & 122.8*** \\
 & Safety    & .060 & .093 & .140 & .080 & 9.6***  & .047 & 123.3*** \\
\bottomrule
\end{tabular}}
\caption{Full structural decomposition by reward model and benchmark split. $\Delta R^2$ is the incremental variance explained; $F$ is the incremental $F$-statistic. Significance: {*}\,$p<0.05$, {**}\,$p<0.01$, {***}\,$p<0.001$. Each split contains $N = 2{,}300$ instances across 23 languages.}
\label{tab:structural_full}
\end{table*}

\subsection{Within-Language Regression Details}
\label{sec:appendix_within_language}

Table~\ref{tab:within_language} summarizes within-language instance-level regressions (fit separately per language) and counts how often within-language slopes are significantly positive vs. negative. 

\begin{table*}[h]
\centering

\small
\begin{tabular}{ll rrr rrr}
\toprule
\textbf{Model} & \textbf{Split} & $\bar{\beta}_w$ & Med.\,$\beta_w$ & $\bar{R}^2$ & \#Sig$^+$ & \#Sig$^-$ & \#NS \\
\midrule
BTRM-Qwen2-7B & Chat & 0.201 & 0.169 & 0.014 & 3 & 0 & 20 \\
 & Chat-Hard & 0.813 & 0.770 & 0.154 & 23 & 0 & 0 \\
 & Reasoning & 0.905 & 0.924 & 0.253 & 23 & 0 & 0 \\
 & Safety & 0.153 & 0.268 & 0.022 & 2 & 1 & 20 \\
\midrule
Skywork-Gemma-27B & Chat & -0.094 & -0.183 & 0.013 & 1 & 0 & 22 \\
 & Chat-Hard & -0.880 & -0.860 & 0.022 & 0 & 3 & 20 \\
 & Reasoning & 1.974 & 1.961 & 0.103 & 23 & 0 & 0 \\
 & Safety & 0.724 & 0.694 & 0.017 & 2 & 0 & 21 \\
\midrule
Skywork-LLaMA-8B & Chat & -0.921 & -0.840 & 0.032 & 0 & 7 & 16 \\
 & Chat-Hard & -1.730 & -1.687 & 0.071 & 0 & 16 & 7 \\
 & Reasoning & 1.616 & 1.654 & 0.058 & 19 & 0 & 4 \\
 & Safety & 1.619 & 1.619 & 0.054 & 16 & 0 & 7 \\
\midrule
URM-LLaMA-8B & Chat & -1.384 & -1.184 & 0.128 & 0 & 16 & 7 \\
 & Chat-Hard & -1.331 & -1.231 & 0.141 & 0 & 23 & 0 \\
 & Reasoning & 0.371 & 0.423 & 0.031 & 8 & 0 & 15 \\
 & Safety & 1.116 & 1.053 & 0.101 & 23 & 0 & 0 \\
\bottomrule
\end{tabular}
\caption{Within-language instance-level regressions: $\mathrm{Score}(x) \sim \beta \cdot \mathrm{NLL}(x) + \varepsilon$, fit separately for each of the 23 languages. $\bar{\beta}_w$ is the mean standardized slope across languages; $\bar{R}^2$ is the mean explained variance; \#Sig$^+$/\#Sig$^-$ count languages with significant positive/negative slopes ($p < 0.05$); \#NS counts non-significant languages.}
\label{tab:within_language}

\end{table*}

\begin{figure}[h]
    \centering
    \includegraphics[width=\linewidth]{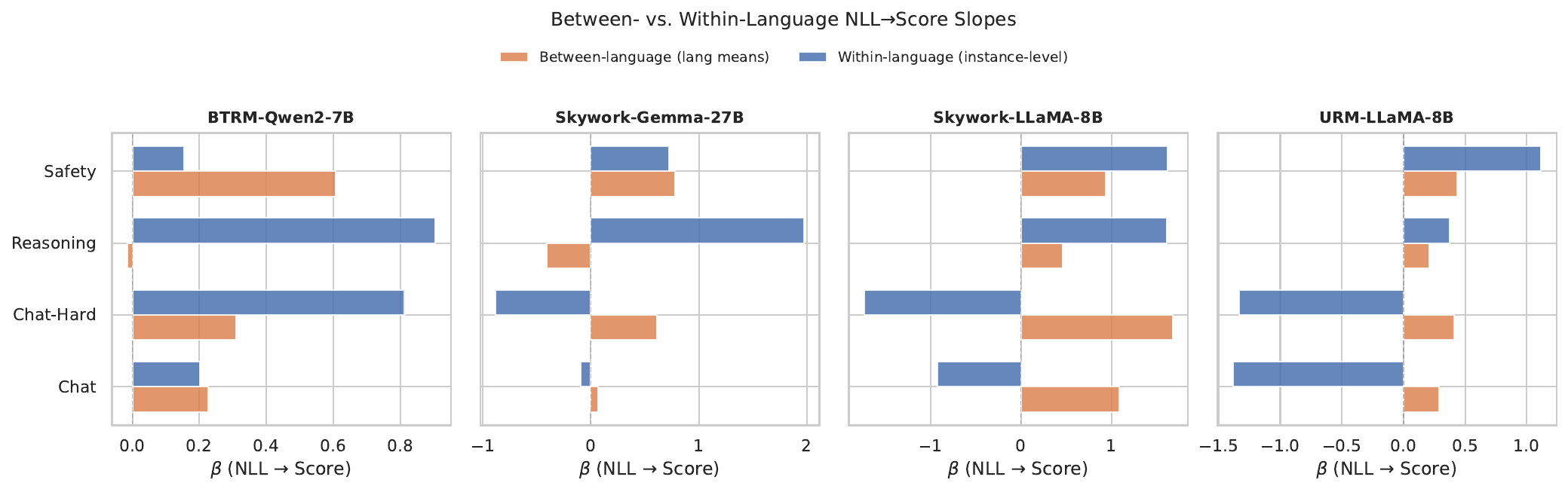}
    \caption{Between-language (orange) vs.\ within-language (blue) NLL$\to$Score slopes for each model and split. A large between-language slope with a near-zero within-language slope indicates a language-level prior; comparable slopes suggest an instance-level difficulty effect.}
    \label{fig:between_within}
\end{figure}

\section{Language-Conditioned Score Priors}
\label{sec:appendix_score_priors}

We evaluate a simple post-hoc correction implied by the additive prior model in Equation~\ref{eq:lang_prior_model}: each reward model shifts scores by a language-specific offset that is constant across items. Because such offsets cancel in paired comparisons, they can preserve pairwise accuracy while creating large cross-language differences under fixed thresholds.

\paragraph{Experiment: additive-bias decomposition.}
For each reward model--benchmark combination, we estimate $b(\ell)$ as the deviation of the per-language mean from the global mean:
\begin{equation}
\label{eq:bias-estimate}
  \hat{b}(\ell) \;=\; \mu_\ell - \mu_{\mathrm{global}},
  \qquad
  \mu_\ell = \frac{1}{|\mathcal{D}_\ell|}\sum_{x \in \mathcal{D}_\ell} s(x,\ell),
\end{equation}
where $\mathcal{D}_\ell$ denotes the set of items evaluated in language $\ell$.
We then subtract $\hat{b}(\ell)$ from every score to obtain $s'(x,\ell) = s(x,\ell) - \hat{b}(\ell)$ and re-evaluate (i)~the cross-language variance of per-language mean scores and (ii)~the acceptance-rate gap under a fixed global median threshold.

\paragraph{Results.}
Table~\ref{tab:additive-bias} reports results for four reward models across all four benchmarks (16 runs total).

\begin{table}[h]
\centering
\small
\vspace{0.5em}
\begin{tabular}{
  l
  S[table-format=3.0]
  S[table-format=2.1]
  S[table-format=2.1]
  S[table-format=2.1]
}
\toprule
\textbf{Benchmark}
  & {\textbf{Var.\,Red.\,(\%)}}
  & {\textbf{Gap\textsubscript{pre}\,(pp)}}
  & {\textbf{Gap\textsubscript{post}\,(pp)}}
  & {\textbf{Gap\,Red.\,(\%)}} \\
\midrule
Chat        & 100 & 33.8 & 13.5 & 55.2 \\
Chat-Hard   & 100 & 43.0 & 13.0 & 66.9 \\
Reasoning   & 100 & 20.1 &  7.5 & 56.9 \\
Safety      & 100 & 36.9 & 12.2 & 64.8 \\
\midrule
\rowcolor{gray!12}
\textit{Average} & 100 & 33.4 & 11.6 & 60.9 \\
\bottomrule
\end{tabular}
\caption{%
  Effect of additive bias correction on cross-language score variance and
  acceptance-rate gap for reward models.
  \textbf{Var.\,Red.}: percentage reduction in cross-language variance of
  per-language mean scores after correction (100\% = fully eliminated).
  \textbf{Gap\textsubscript{pre}} and \textbf{Gap\textsubscript{post}}:
  acceptance-rate gap (max $-$ min across languages, in percentage points)
  at the median threshold, before and after correction.
  \textbf{Gap\,Red.}: percentage reduction in acceptance-rate gap.
}
\label{tab:additive-bias}
\end{table}

Two findings stand out. First, the additive correction eliminates 100\% of the cross-language variance in \emph{every} run, confirming that differences in per-language mean scores are perfectly captured by a single scalar offset $\hat{b}(\ell)$ per language. Second, the acceptance-rate gap drops by 60.9\% on average (from 33.4\,pp to 11.6\,pp), with the largest reduction on Chat-Hard (66.9\%) and Safety (64.8\%). The residual gap of 11.6\,pp suggests that while the additive component is dominant, higher-order effects (e.g., language--item interactions) also contribute to cross-language variation.

\paragraph{What the additive correction cannot fix.}
The residual has structure. First, languages differ not only in mean but in \emph{dispersion}: even after centering, per-item score spread is widest for English and narrowest for Persian, Turkish, and Hindi (Table~\ref{tab:residual_structure}, left). A threshold shift cannot equalize acceptance rates between distributions of different shape. Second, we decompose each centered score into a shared item-difficulty term and a language$\times$item interaction: writing $d(x)$ for the per-item mean of centered scores across languages, $d(x)$ and the remainder are orthogonal, so their variances add. Shared item difficulty explains 81.2\% of centered-score variance on average, leaving 18.8\% to language$\times$item interaction (Table~\ref{tab:residual_structure}, right): some items are scored differently depending on their language, which no per-language scalar can correct.

\begin{table}[h]
\centering
\small
\begin{minipage}[t]{0.42\textwidth}
\centering
\begin{tabular}{lc}
\toprule
\textbf{Language} & \textbf{Score std} \\
\midrule
Persian    & 4.69 \\
Turkish    & 4.70 \\
Hindi      & 4.70 \\
\textit{(17 languages)} & 4.88--5.39 \\
Portuguese & 5.44 \\
German     & 5.44 \\
English    & 5.92 \\
\bottomrule
\end{tabular}
\end{minipage}\hfill
\begin{minipage}[t]{0.54\textwidth}
\centering
\begin{tabular}{lcc}
\toprule
\textbf{Reward model} & \textbf{Shared item} & \textbf{Lang.\,$\times$\,item} \\
\midrule
BTRM-Qwen     & 89.2\% & 10.8\% \\
URM-LLaMA     & 75.6\% & 24.4\% \\
Skywork-Gemma & 74.7\% & 25.3\% \\
Skywork-LLaMA & 85.2\% & 14.8\% \\
\midrule
\textit{Average} & 81.2\% & 18.8\% \\
\bottomrule
\end{tabular}
\end{minipage}
\caption{\textbf{Structure of the residual after additive correction.} \textbf{Left:} per-language standard deviation of centered reward scores; the three narrowest and three widest of the 23 languages are shown. \textbf{Right:} share of centered-score variance explained by shared item difficulty versus language$\times$item interaction, per reward model, averaged across the four benchmark splits.}
\label{tab:residual_structure}
\end{table}

\paragraph{Language-specific thresholds and the LID attack surface.}
Subtracting $\hat{b}(\ell)$ and applying a global threshold $T$ is mathematically equivalent to keeping raw scores and applying a language-specific threshold $T_\ell = T + \hat{b}(\ell)$, so the correction above is also an evaluation of language-specific thresholding. Global thresholds nevertheless remain the deployment default: RLHF applies a single scalar reward function across multilingual rollouts~\citep{NEURIPS2022_b1efde53}, reward-model benchmarks report one aggregate accuracy, and deployed safety classifiers apply one decision boundary across languages~\citep{51952}, with measurable consequences; Perspective API has been shown to moderate roughly four times more German tweets than their English translations~\citep{nogara2024toxicbiasperspectiveapi}.

Language-specific thresholding is thus a practical takeaway of this work, but it is not free: it requires per-language calibration data and a reliable language identification (LID) step, both fragile for low-resource and code-switched inputs~\citep{mendels-etal-2018-collecting, dogruoz-etal-2021-survey, suarez-etal-2026-commonlid}. \S\ref{sec:pairwise_bias} and Table~\ref{tab:code_switch} demonstrate the resulting attack surface: code-switched prompts that defeat the LID step are judged against the wrong threshold, raising acceptance from the calibrated 50\% to 75\%.

\begin{table}[h]
\centering
\small
\begin{tabular}{llcc}
\toprule
\textbf{Setup} & \textbf{LID label} & \textbf{Threshold applied} & \textbf{Acceptance rate} \\
\midrule
Pure English (Safety) & en & $T_{\mathrm{en}} = -22.06$ & 50.0\% \\
Pure Hindi (Safety)   & hi & $T_{\mathrm{hi}} = -13.45$ & 50.0\% \\
Code-switched: en wrapper + hi content & en (mislabel) & $T_{\mathrm{en}} = -22.06$ & 75.0\% \\
\bottomrule
\end{tabular}
\caption{\textbf{Code-switching defeats language-specific thresholds.} Skywork-Reward-LLaMA-3.1-8B-v0.2 on \textsc{M-RewardBench} Safety. Thresholds are the 50th percentile of each pure language, so both pure settings accept 50\% by construction; mis-identified code-switched prompts are judged against the lenient English threshold.}
\label{tab:code_switch}
\end{table}

\section{Evaluation Prompts}
\label{sec:appendix_prompts}

For reproducibility, we list below the exact prompts used for both evaluation paradigms.

\subsection{Prompted \textsc{LLM-as-a-Judge}}

Each prompted evaluator receives a \textbf{system prompt} and a \textbf{user prompt} in the target evaluation language. The user prompt contains two placeholders: \texttt{\{prompt\}} (the instruction being evaluated) and \texttt{\{answer\}} (the response being evaluated). All 22 non-English prompts follow the same template structure, professionally translated into each target language. Below we show the English template.

\paragraph{System prompt.}
\begin{small}
\begin{verbatim}
You are a fair judge assistant tasked with providing clear, objective
feedback, ensuring each assessment reflects the absolute standards set
for performance. You will output an absolute quality score from 1 to 5.
\end{verbatim}
\end{small}

\paragraph{User prompt.}
\begin{small}
\begin{verbatim}
###Task Description:
An instruction and a response to evaluate are given.
1. Output a score that is an integer between 1 and 5.
2. The output format should look as: "SCORE: (an integer number between 1 and 5)"
3. Please do not generate any other opening, closing, or explanations.

###The instruction to evaluate:
{prompt}

###Response to evaluate:
{answer}
\end{verbatim}
\end{small}

\noindent For each of the remaining 22 languages (ar, cs, de, el, es, fa, fr, he, hi, id, it, ja, ko, nl, pl, pt, ro, ru, tr, uk, vi, zh), both the system prompt and user prompt are translated into the target language while preserving the identical structure, scoring scale (1--5), output format requirement (\texttt{SCORE: ...}), and placeholder variables (\texttt{\{prompt\}}, \texttt{\{answer\}}).

\subsection{Trained Reward Models}

Trained reward models do not receive an evaluation rubric. Instead, the instruction--response pair is formatted as a standard two-turn conversation:
\begin{small}
\begin{verbatim}
[
  {"role": "user",      "content": "<instruction text>"},
  {"role": "assistant", "content": "<response text>"}
]
\end{verbatim}
\end{small}

\noindent The model's scalar reward head then produces a single continuous score for the given conversation. No system prompt, evaluation rubric, or score format is specified; the reward model scores the response purely based on its learned preference function.

\section*{Use of Large Language Models}
LLMs were used as auxiliary tools to assist with code generation and debugging, and to polish the writing and presentation of the manuscript. All scientific contributions, experimental decisions, and interpretations were made by the authors.

\end{document}